\documentclass{article}     

\usepackage{arxiv}
\usepackage[utf8]{inputenc} 
\usepackage[T1]{fontenc}    
\usepackage{hyperref}       
\usepackage{url}            
\usepackage{booktabs}       
\usepackage{amsfonts}       
\usepackage{nicefrac}       
\usepackage{microtype}      
\usepackage{lipsum}
\usepackage{graphicx}
\usepackage{caption}
\usepackage{subcaption}
\usepackage{multirow}
\usepackage{amsmath,amsfonts}
\usepackage{tensor}
\usepackage{multirow,multicol}
\usepackage{orcidlink}
\usepackage[capitalise]{cleveref}
\usepackage{wrapfig}
\title{A Biomimetic Vertebraic Soft Robotic Tail for High-Speed, High-Force Dynamic Maneuvering}
\author{
    Sicong Liu \\
    Shenzhen Technology University\\
    \And
    Nan Huang, Jianhui Liu, Fang Chen, Wenjian Yang \\
    Southern University of Science and Technology\\
    \AND
    Yukang Nie, Yipan Zhu \\
    Shenzhen Technology University\\
    \And
    Juan Yi \\
    Great Bay University\\
    \And
    Yu Zheng \\
    Tencent Robotics X\\
    \And
    Zheng Wang \\
    Wisson Robotics\\
    \AND
    Wanchao Chi\(^{*}\) \\
    Tencent Robotics X\\
    \texttt{wanchao.chi@gmail.com} \\
    \And
    Chaoyang Song\thanks{This work was partly supported by the National Natural Science Foundation of China under Grants 52475302 and 62473189, the Science, Technology, and Innovation Commission of Shenzhen Municipality under Grant JCYJ20220530114615034 and JCYJ20220818100417038, and Guangdong Basic and Applied Basic Research Foundation Grant 2021A1515110658. (\textit{Corresponding authors: Wanchao Chi and Chaoyang Song})} \\
    asRobotics\\
    \texttt{songcy@ieee.org} \\
}
\begin{document}
\maketitle
\begin{abstract}

    Robotic tails can enhance the stability and maneuverability of mobile robots, but current designs face a trade-off between the power of rigid systems and the safety of soft ones. Rigid tails generate large inertial effects but pose risks in unstructured environments, while soft tails lack sufficient speed and force. We present a Biomimetic Vertebraic Soft Robotic (BVSR) tail that resolves this challenge through a compliant pneumatic body reinforced by a passively jointed vertebral column inspired by musculoskeletal structures. This hybrid design decouples load-bearing and actuation, enabling high-pressure actuation (up to 6 bar) for superior dynamics while preserving compliance. A dedicated kinematic and dynamic model incorporating vertebral constraints is developed and validated experimentally. The BVSR tail achieves angular velocities above 670°/s and generates inertial forces and torques up to 23.6 N and 2.48 Nm with a 500 g tip payload, indicating over 200\% improvement compared to non-vertebraic designs. Demonstrations of rapid cart stabilization, obstacle negotiation, high-speed steering, and quadruped integration confirm its versatility and potential for application in agile robotic platforms.
    
\end{abstract}
\section{Introduction}
\label{sec:Intro}

    The tail is a masterful evolutionary solution for dynamic locomotion, enabling animals to achieve remarkable feats of stability and agility \cite{hickman1979the}. The functional utility of this appendage is rooted in the principles of classical mechanics, particularly the conservation of angular momentum. This is often accomplished through a process known as \textit{inertial adjustment}, where rapid, controlled movements of the tail generate reaction forces and torques that are imparted onto the main body to regulate its orientation and momentum in real time. For instance, cheetahs, during their high-speed bounding gait, use their long, muscular tails as aerodynamic rudders and inertial counterweights to modulate yaw and roll, allowing for exceptionally sharp turns and stable braking \cite{patel2014rapid, patel2013rapid}. In the aerial realm, animals like geckos and lizards execute rapid mid-air self-righting maneuvers by swinging their tails, inducing a counter-rotation in their bodies to ensure a safe landing orientation \cite{jusufi2008active, libbytail}. Even during terrestrial or arboreal locomotion, animals from kangaroos to squirrels leverage their tails for a spectrum of dynamic tasks, from providing a ``fifth leg'' for postural stability to recovering from unexpected falls and slips \cite{alexander1975the, fukushima2021inertial}. These biological archetypes, which demonstrate a sophisticated functional integration of sensing, neural control, and musculoskeletal actuation, provide a rich foundation for designing robotic tails that augment the dynamic performance of mobile robots \cite{machairas2015on, briggs2012tails}.
    
    Inspired by these natural mechanisms, roboticists have long sought to replicate their function to enhance the agility and robustness of mobile systems. Early pioneering work included the Uniroo, a monopedal hopping robot that employed a simple tail for pitching stabilization \cite{zeglin1991uniroo}. Subsequent research in this domain has primarily focused on rigid, articulated tails that typically operate as pendulum-like mechanisms. These designs range from single-degree-of-freedom (DOF) systems for planar regulation in the pitch, yaw, or roll axes \cite{patel2014rapid, liu2014a, wang2021balance, kohut2012effect, berenguer2008zappa, patel2013rapid, changsiu2011a, de2015the, Wang2021Volumetrically}, to 2-DOF mechanisms that provide more generalized spatial control \cite{changsiu2013a, patel2015on}. The high stiffness and well-defined kinematics of these rigid systems enable them to generate significant, predictable inertial effects at high speeds. However, their fundamental design presents a critical set of trade-offs. The primary limitation is their lack of compliance, which limits their utility in unstructured environments and poses significant safety risks during physical human-robot interaction \cite{laschi2014soft}.       
    
    To address the inherent safety and compliance limitations of rigid systems, the field of soft robotics presents a compelling alternative paradigm. The intrinsic compliance of soft robots, derived from their deformable materials and structures, offers inherent safety, adaptability to uncertain environments, and robustness to physical impacts \cite{laschi2016soft, Yang2020Scalable, xie2023octopus, Wu2024Vision, godage2012pneumatic}. This has motivated the development of a new class of soft robotic tails, including the soft prehensile tail with grasping and dynamic mobility \cite{XM2025softtail}, aquatic robots with flexible tails that emulate fish locomotion for propulsion and maneuvering \cite{marchese2014autonomous}, and even the novel wearable tail for human balance assistance \cite{nabeshima2019arque}. However, while these pioneering systems validate the potential of soft appendages, they also reveal a persistent performance gap. This deficit is rooted in the fundamental properties of the soft materials themselves, which typically exhibit low stiffness and significant viscoelasticity, making it challenging to generate sufficient force at high frequencies. The low actuation authority of most soft systems, combined with the immense difficulty of accurately modeling and controlling their near-infinite degrees of freedom, limits their ability to produce the rapid, high-magnitude accelerations necessary for effective dynamic regulation of a large robotic platform. A family of innovative compliant and continuum robotic tails provides a feasible paradigm to achieve spatial loading for the adjustment of robots, including the hyper-redundant continuum tail for inertial adjustment \cite{rone2014continuum}, the cable-driven tail system capable of spatial motion to generate spatial loading \cite{saab2019design}, and the flexible universal spatial robotic tail \cite{rone2018design} utilizing backbone architectures. Because the rigid components occupy the outer surfaces of the compliant tails, they are not intended to operate in environments with unexpected human or object interactions. Consequently, the field faces a critical unmet need: a robotic appendage that combines the raw inertial authority of rigid or compliant systems, such as high-speed and high-force output, with the inherent safety and adaptability of soft structures. The absence of such a system currently precludes the deployment of agile, dynamic robots in unstructured, human-centric environments.

    \begin{figure*}[!t]
        \centering
        \includegraphics[width=1\linewidth]{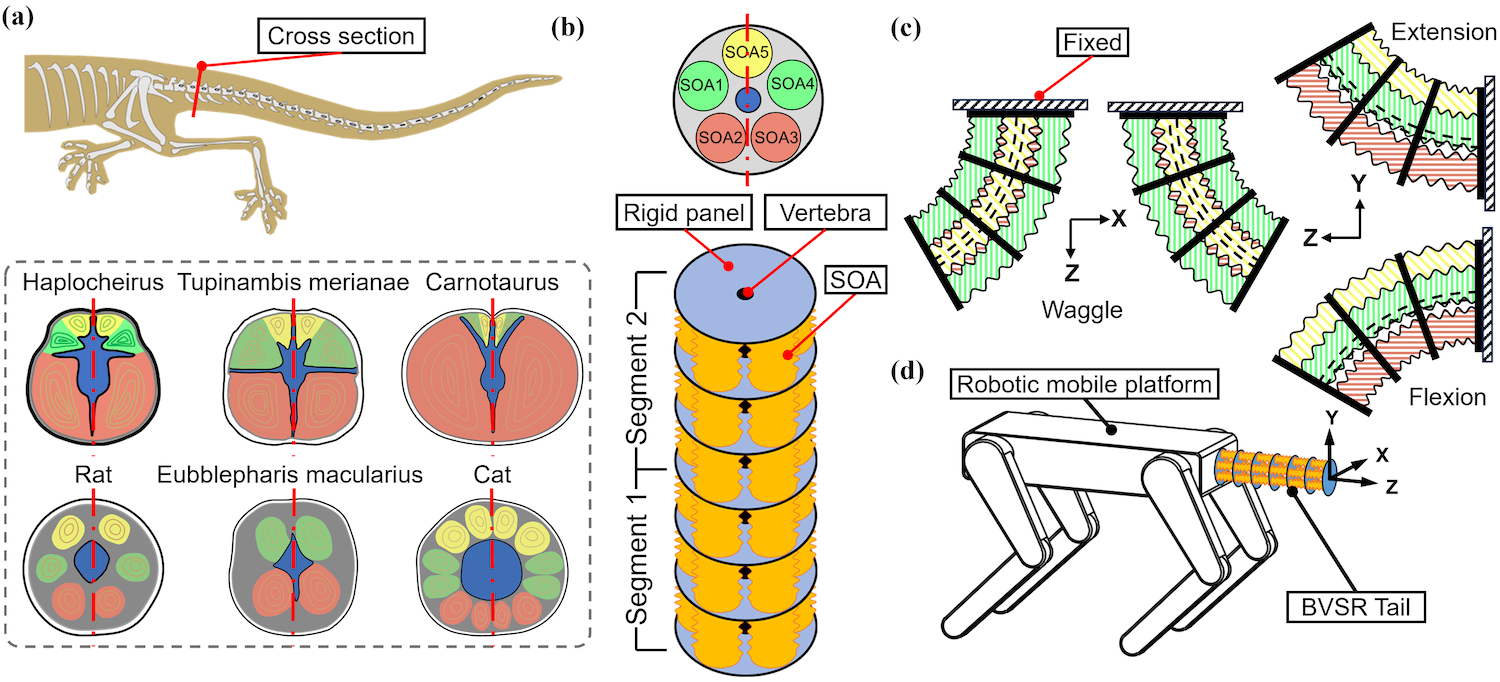}
        \caption{
            \textbf{The Biomimetic Vertebraic Soft Robotic (BVSR) tail proposed in this study.}
            (a) The BVSR tail consists of 2 independently controlled 2-degree-of-freedom (DOF) segments, each composed of 3 joints.
            (b) The biomimicking layout of actuators and elastic rod as vertebra inspired by the alvarezsaurian tail (Yellow: M. spinalis; Green: M. longissimus; Light red: M. caudofemoralis longus; Blue: caudal vertebra).
            (c) The definition of the 2-DoF BVSR segment motions. 
            (d) The BVSR tail is installed on a robotic mobile platform (e.g., a quadruped robot).
        }
        \label{fig:1_Overview}
    \end{figure*}
    
    This paper introduces a Biomimetic Vertebraic Soft Robotic (BVSR) tail, as shown in Fig.~\ref{fig:1_Overview}, specifically designed to address the trade-offs above by establishing a novel hybrid design approach. Our method involves the functional integration of a compliant pneumatic body with an internal, passively jointed vertebral column. This central element performs a crucial dual role: it acts as a structural backbone that bears the tensile loads from high-pressure actuation, allowing the system to generate large forces without material failure, and it serves as a kinematic constraint that reduces the complex, high-dimensional deformation of the soft body into a predictable, low-dimensional bending motion. This architectural choice makes the system's modeling and control more tractable while enabling a level of dynamic performance previously unattainable in soft robotic appendages. The primary contributions of this work are threefold:
    \begin{itemize}
        \item The formulation and physical realization of a vertebraic soft robotic design principle, wherein a passive kinematic constraint enables high-pressure (6 bar) actuation in an otherwise compliant structure to achieve superior angular velocity and inertial output.
        \item A comprehensive Euler-Lagrange dynamic model that, among the first in the pneumatic soft robotic tail with a continuous vertebra, explicitly incorporates the kinematic constraints imposed by an internal vertebral structure and captures the dominant experimental trends.
        \item Rigorous experimental validation of the tail's performance envelope and a demonstration of its functional efficacy in dynamic tasks, including inertial assistance for a wheeled mobile robot and successful integration with quadrupedal platforms, validating its practicality and versatility.
    \end{itemize}
    
    The remainder of this paper is organized as follows. Section~\ref{sec:Concept} describes the concept and modeling of the BVSR tail. Section~\ref{sec:Implement} presents its physical implementation and characterization. Section~\ref{sec:Experiments} details the experimental validation and functional demonstrations. Section~\ref{sec:Discussion} discusses the implications of our findings, limitations, and comparisons to the state of the art. Finally, Section~\ref{sec:Conclusion} concludes the paper and discusses future work.   

\section{Concept and Modeling of the BVSR Tail}
\label{sec:Concept}

\subsection{Design Concept of the BVSR Tail}
    
    The agility of vertebrate animals often relies on the tail acting as a dynamic stabilizer during rapid maneuvers like predation or evasion \cite{ostrom2019osteology}. This principle of inertial stabilization is highly desirable for enhancing the performance of mobile robotic platforms. Our design for the BVSR tail is inspired by the potent musculoskeletal architecture found in the tails of agile reptiles and mammals (e.g., Alvarezsauria, felids, and murids) \cite{mesotail, norby2021enabling, persons2011dinosaur, payne2017blood, wada1994anatomical, vanhoutte2002in}. As illustrated in Fig.~\ref{fig:1_Overview}, the BVSR tail translates these biological principles into a novel robotic mechanism.
    
    The biological tail's muscular arrangement fundamentally dictates its motion capabilities. In many vertebrates, the tail muscles are arranged symmetrically around the vertebral column. For example, the alvarezsaurian tail features six primary muscle bundles (M. spinalis, M. longissimus, M. caudofemoralis longus) arranged in bilateral symmetry \cite{mesotail}. Inspired by this functional layout, we arranged five Soft Oscillating Actuators (SOAs) in a planar, mirror-symmetric configuration, as shown in Fig.~\ref{fig:1_Overview}(b). This arrangement mimics the distinct muscle groups responsible for flexion, extension, and wagging, and contrasts with rotationally symmetric designs, such as the 3-DOF TESV joint \cite{liu2021vertebraic}, providing anisotropic bending characteristics tailored for inertial stabilization.
    
    The central component of our design is a continuous elastic rod that functions as a synthetic \textit{vertebra}. This element serves as a critical \textit{kinematic constraint}, resisting axial elongation while permitting controlled bending. Its primary role is to convert the translational expansion of the SOAs into structured, predictable 2-DOF rotations of the entire joint. This design is also highly tunable; by selecting rods of different materials, geometries, or cross-sections, the tail's passive stiffness and dynamic response can be customized to suit specific robotic platforms and operational demands.
    
\subsection{Kinematic Modeling of the BVSR Joint}
    
\begin{figure}[b]
        \centering
        \includegraphics[width=0.5\linewidth]{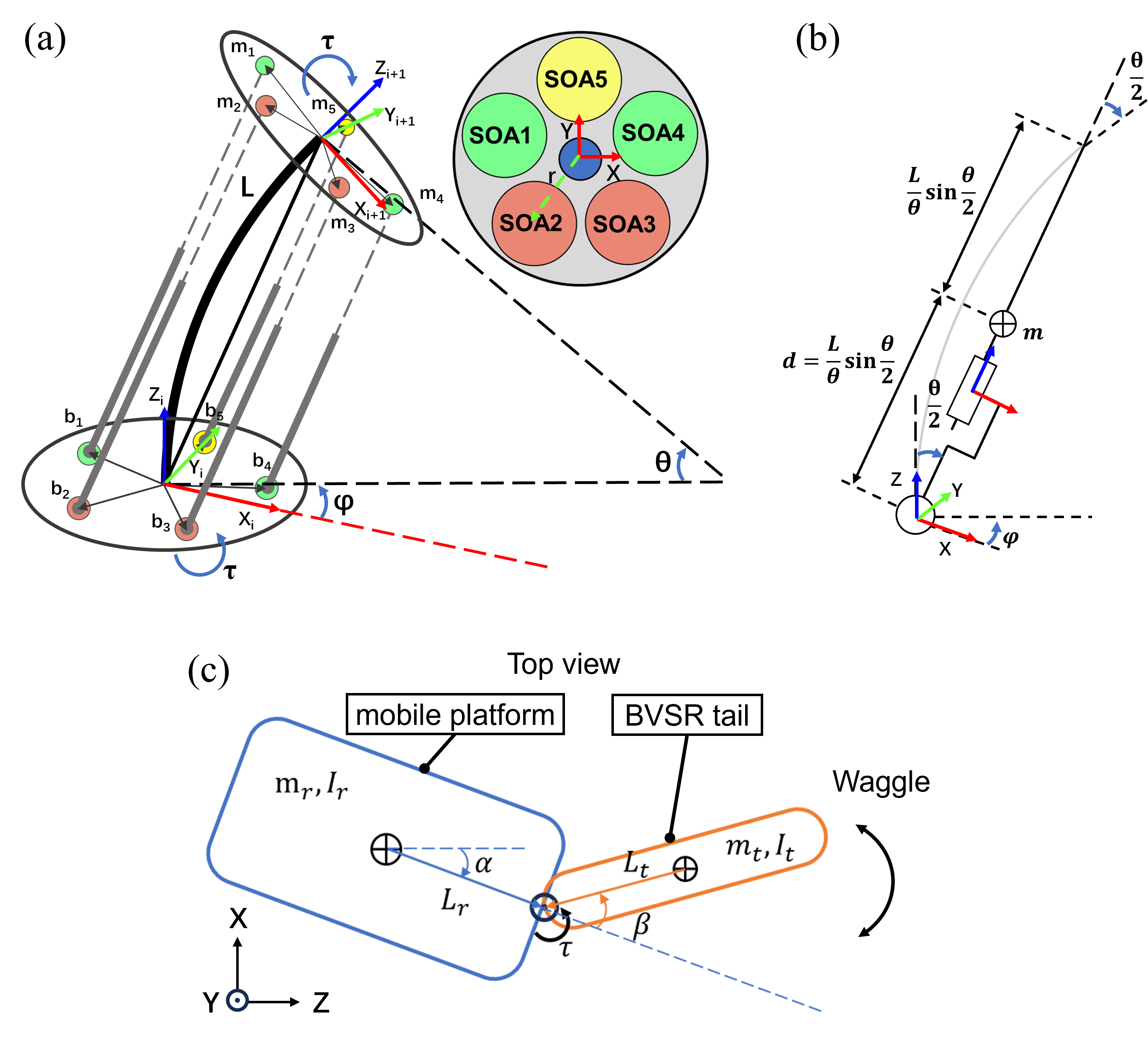}
        \caption{
            \textbf{Modeling schematics.}
            (a) Kinematics of a single BVSR joint with each SOA as a prismatic joint.
            (b) Simplified dynamic model as a pendulum. 
            (c) Planar two-body model of the tailed mobile platform.
        }
        \label{fig:2_JointKinematics}
    \end{figure}

    Figure~\ref{fig:2_JointKinematics}(a) shows our kinematic model of a single BVSR joint by assuming its motion follows a \textit{constant curvature} profile, a common and effective simplification for continuum-style robots. This assumption, while a simplification for general continuum robots, is physically enforced in our design by the central vertebra, which acts as a non-extensible elastic backbone and constrains the soft body's complex deformation to a predictable, planar arc of constant length \( L \) during bending. The joint's configuration can thus be described by a bending angle \( \theta \) and a bending plane orientation \( \varphi \).

    The length \( l_i \) of the \textit{i-th} SOA is a function of the joint's configuration, i.e. \( l_{i} = \left\Vert  m_{i} - b_{i} \right\Vert \), where \( m_{i} \) denotes the homogeneous coordinates of the center of the \textit{i-th} SOA's end face on the moving platform in the base frame, and \( b_{i} \) denotes the center on the \textit{i-th} SOA's base panel in the base frame. Considering the homogeneous transformation matrix \( \mathbf{T}_{s} \) between the moving and base platforms of the joint, the kinematic model is presented as follows,
    \begin{equation} \label{eq:1}
        l_{i} = \left\Vert \mathbf{T}_{\mathbf{s}} b_{i} - b_{i} \right\Vert,
    \end{equation}
    Based on the constant curvature assumption, \( \mathbf{T}_{s} \) is defined as:
    \begin{equation} \label{eq:2}
        \begin{matrix}
            \mathbf{T}_{s} = \\
            \begin{bmatrix}
            c_{\varphi}^{2} c_{\theta} + s_{\varphi}^{2} & c_{\varphi}s_{\varphi}(c_{\theta} -1) & c_{\varphi} s_{\theta} & \frac{L c_{\varphi} (1 - c_{\theta} )}{\theta}  \\
            c_{\varphi}s_{\varphi}( c_{\theta} - 1) & s_{\varphi}^{2} c_{\theta} + c_{\varphi}^{2} & s_{\varphi} s_{\theta} & \frac{L s_{\varphi} (1 - c_{\theta})}{\theta}  \\ 
            -c_{\varphi} s_{\theta} & -s_{\varphi} s_{\theta} & c_{\theta} & \frac{L s_{\theta}}{\theta} \\ 
            0 & 0 & 0 & 1 \\ 
            \end{bmatrix},
        \end{matrix}
    \end{equation}
    where \( c_{(\cdot)} = \cos(\cdot) \) and \( s_{(\cdot)} = \sin(\cdot) \). The static output force \( F_i \) of an SOA is modeled as a function of the input pressure \( P_i \) and its change in length \( \Delta l_i = l_i - L \), following the model from \cite{su2020a}:
    \begin{equation} \label{eq:3}
        F_{i} = P_{i} A - k \Delta l_{i},
    \end{equation}
    where \(A\) is the effective cross-sectional area of the SOA and \(k\) is a pressure-dependent stiffness coefficient:
    \begin{equation} \label{eq:4}
        k = m_{a} P_{i} + n_{a}.
    \end{equation}
    The empirical parameters \(m_{a}\) and \(n_{a}\) are determined through characterization of a single SOA. By combining the kinematic and static force models, we obtain the inverse static mapping from a desired joint configuration (\( \theta, \varphi \)) and external force \(F_i\) to the required input pressure \( P_i \):
    \begin{equation} \label{eq:5}
        P_{i} =\frac{F_{i} + n_{a}(l_i - L)}{A - m_{a}(l_i -L)}, \quad \text{where } l_i = l_i(\theta, \varphi).
    \end{equation}
    
\subsection{Dynamic Modeling of the BVSR Joint}
    
    The dynamics of the BVSR joint are derived using the Euler-Lagrange formulation. The joint is simplified to a rigid body with its Center of Mass (CoM) located at a distance \( d =\frac{L}{\theta} s_{\frac{\theta}{2}} \) from the base, as shown in Fig.~\ref{fig:2_JointKinematics}(b). The equations of motion for the generalized coordinates \( \mathbf{q} =[\varphi, \theta]^{T} \) are given by:
    \begin{equation} \label{eq:6}
        \mathbf{M}(\mathbf{q}) \ddot{\mathbf{q}} + \mathbf{c}(\mathbf{q}, \dot{\mathbf{q}})\dot{\mathbf{q}}  + \mathbf{D}\dot{\mathbf{q}} + \mathbf{K}(\mathbf{q}) + \mathbf{G}(\mathbf{q}) = \boldsymbol{\tau}. 
    \end{equation}
    In Eq.~\eqref{eq:6}, the term $\mathbf{M}(\mathbf{q}) \ddot{\mathbf{q}}$ represents the inertial forces of the joint, where the inertia matrix $\mathbf{M}(\mathbf{q})$ is configuration-dependent. The vector $\mathbf{c}(\mathbf{q}, \dot{\mathbf{q}})$ captures the velocity-dependent Coriolis and centrifugal forces, which become significant during high-speed maneuvers. The terms $\mathbf{D}\dot{\mathbf{q}}$, $\mathbf{K}(\mathbf{q})$, and $\mathbf{G}(\mathbf{q})$ represent the dissipative damping, passive elastic restoring forces from the vertebra and actuators, and gravitational loads, respectively. The generalized torque $\boldsymbol{\tau}$ is the input to the system, derived from the pneumatic actuation.
    
    The inertia matrix \( \mathbf{M}(\mathbf{q}) \) is:
    \begin{equation} \label{eq:7}
        \begin{matrix}
            \mathbf{M}(\mathbf{q}) =  \\
            \begin{bmatrix}
                I_{\varphi}+  \frac{m L^{2}}{\theta^{2}}s^2_{\frac{\theta}{2}}  & 0 \\ 
                0 & \frac{I_{\theta} }{4} + \frac{m L^{2}}{4\theta^{4}} (\theta^2-2\theta s_{\theta}-2c_{\theta}+2)
            \end{bmatrix}.
        \end{matrix}
    \end{equation}
    The Coriolis/centrifugal vector \( \mathbf{c}(\mathbf{q}, \dot{\mathbf{q}}) \) and gravity vector \( \mathbf{G}(\mathbf{q}) \) are:
    \begin{equation} \label{eq:8}
        \begin{matrix}
            \mathbf{c}(\mathbf{q}, \dot{\mathbf{q}}) = \\
            \begin{bmatrix}
                \frac{m L^{2}}{\theta^{3}}( \frac{\theta}{2} s_{\theta} -2s^2_\frac{\theta}{2}) \dot{\varphi} \dot{\theta} \\ 
                \frac{m L^{2}}{\theta^{5}}( \theta s_{\theta} -2s^2_\frac{\theta}{2}-\frac{\theta^2}{2}c^2_\frac{\theta}{2}) \dot{\theta}^{2} - \frac{m L^{2}}{2\theta^{3}}( \frac{\theta}{2} s_{\theta} -2s^2_\frac{\theta}{2}) \dot{\varphi}^2 \\
            \end{bmatrix},
        \end{matrix}
    \end{equation}    
    \begin{equation} \label{eq:9}
        \mathbf{G}(\mathbf{q}) =  
        \begin{bmatrix}
            0 \\ 
            \frac{mgL}{\theta^{2}}(\frac{3\theta}{2} c_\frac{\theta}{2}c_\theta-s_\frac{\theta}{2}c_\theta- \theta c_\frac{\theta}{2})\\ 
        \end{bmatrix}.
    \end{equation}
    
    The stiffness term \( \mathbf{K}(\mathbf{q}) \) combines the output moment from the SOAs (\( \boldsymbol{\tau}_{a} \)) and the restoring moment of the vertebra (\( \boldsymbol{\tau}_{v} \)), i.e. \( \mathbf{K}(\mathbf{q}) =\boldsymbol{\tau}_{a}- \boldsymbol{\tau}_{v}\). The moment from the actuators is \( \boldsymbol{\tau}_{a} = \mathbf{J}^{T}_{a}\mathbf{F}r \), where \( \mathbf{J}_\mathbf{a} = \frac{\partial q}{\partial l}\in\mathbb{R}^{5 \times 2} \), and  \( r = \frac{D_a}{2}\), \( \mathbf{F} \) is the vector of SOA forces and \( \mathbf{J}_{a} \) is the actuator Jacobian mapping actuator lengths to joint space. The moment from the vertebra is 
    \(
    \boldsymbol{\tau}_{v}
    =
    \begin{bmatrix}
    0,
    \dfrac{EI_p}{L}\theta
    \end{bmatrix}^{T}
    \)
    where \(E\) is the Young's modulus of the rod and \(I_p\) is the moment of inertia. Therefore,  
    \begin{equation} \label{eq:10}
        \mathbf{K}(\mathbf{q}) =\mathbf{J}^{T}_\mathbf{a}\mathbf{P}Ar-\mathbf{J}^{T}_\mathbf{a}kr \boldsymbol{\Delta} \mathbf{l}(\varphi,\theta)-
        \begin{bmatrix}
            0\\
            \dfrac{EI_p}{L}\theta
        \end{bmatrix}.
    \end{equation}
    Considering the dissipative nature of the SOAs, we introduce a linear damper for each SOA. The damper generates a force denoted as  \( f = -c\dot{l_i} \), where \( c \) is the damping coefficient. Similarly, the moment of the dampers is described as 
    \begin{equation} \label{eq:11}
        \boldsymbol{\tau}_{f}= -cr\mathbf{J}_\mathbf{a}\frac{\partial l_i\left(\mathbf{q}\right)}{\partial\mathbf{q}}\dot{\mathbf{q}},
    \end{equation}    
    and the dissipative term  \( \mathbf{D} \) as 
    \begin{equation} \label{eq:12}
        \mathbf{D}= -cr\mathbf{J}_\mathbf{a}\frac{\partial l_i\left(\mathbf{q}\right)}{\partial\mathbf{q}}.
    \end{equation}
    
\subsection{Modeling of the Tailed Mobile Platform}
    
    To analyze the tail's effect on a mobile robot, we model the combined system as a two-body planar mechanism, as shown in Fig.~\ref{fig:2_JointKinematics}(c). The dynamic model is simplified to focus on the case in which the mobile platform outputs only rotational motion. Although the physical tail consists of two segments, the model assumes they bend synchronously. Under this assumption, the tail is represented as an equivalent single continuum structure to obtain a more concise analytical model. This assumption corresponds to the ideal case in which both segments bend cooperatively, resulting in the maximum overall tail deflection. The robot body (mass \(m_r\), inertia \(I_r\)) and the tail (mass \(m_t\), inertia \(I_t\)) are hinged, with their CoMs at distances \(L_r\) and \(L_t\) from the hinge, respectively. The system's orientation is described by the platform angle \( \alpha \), and the relative tail angle \( \beta \).
    
    The Lagrangian \(\mathcal{L} \) for this system can be expressed as the difference between kinetic energy \(T\) and potential energy \(U\), \( \mathcal{L} = T - U \), which can be calculated as:    
    \begin{multline} \label{eq:20}
        \mathcal{L} = \frac{1}{2} m_{t} L_{t}^{2}(\dot{\beta}-\dot{\alpha}  )^{2}+\frac{1}{2} m_{r} L_{r}^{2} \dot{\alpha}^{2} + \frac{1}{2} I_{t} (\dot{\beta}-\dot{\alpha}  )^{2}+  \frac{1}{2}I_{r}\dot{\alpha}^{2}
        \\
        - m_{r} g L_{r} \sin\alpha - m_{t}gL_{t} \sin(\beta-\alpha).
    \end{multline}
    Applying the Euler-Lagrange equations yields the system's dynamics. The equation for the tail's motion relative to the body is:
    \begin{equation} \label{eq:21}
        (I_{t} + m_{t} L_{t}^{2})(\ddot{\beta}-\ddot{\alpha}) + m_{t} g L_{t} \cos( \beta-\alpha ) = \tau,
    \end{equation}
    where \( \tau \) is the inertial torque generated by the tail's actuators. The equation for the body's motion, which is influenced by the tail's inertial coupling, is:
    \begin{equation} \label{eq:22}
        \begin{aligned}
            & (I_{r} + m_{r}L_{r}^{2} + I_{t} + m_{t}L_{t}^{2}) \ddot{\alpha} - (I_{t} + m_{t}L_{t}^{2})\ddot{\beta} \\
            & \qquad + m_{r} g L_{r} \cos\alpha - m_{t} g L_{t} \cos(\beta-\alpha) = 0.
        \end{aligned}
    \end{equation}

\section{System Design and Implementation}
\label{sec:Implement}

\subsection{Hierarchical Design of the BVSR Tail}

\begin{figure*}[htbp]
        \centering
        \includegraphics[width=1\linewidth]{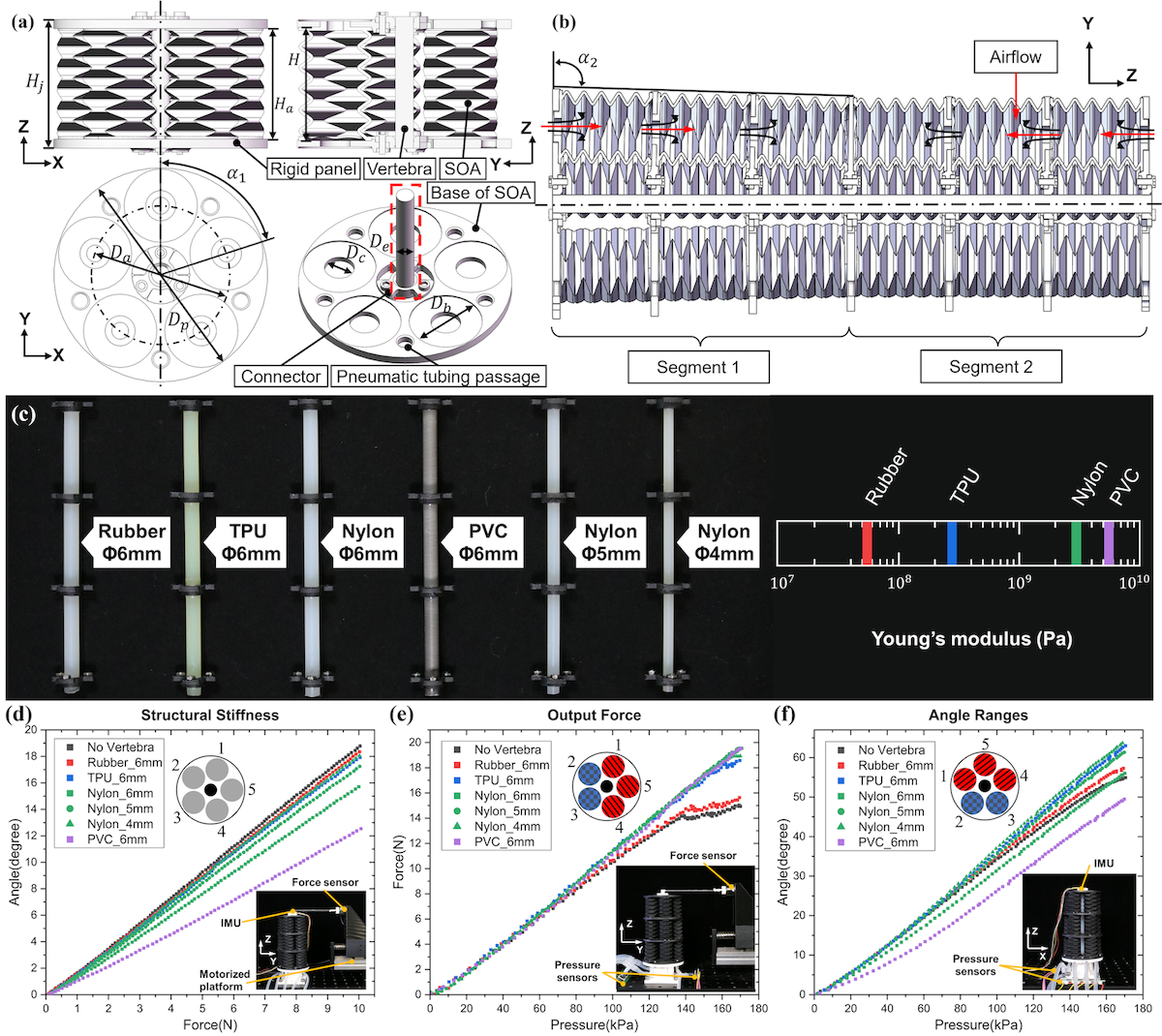}
        \caption{
            \textbf{Design and characterization of the BVSR tail.}
            (a) Geometric parameters of a single joint. 
            (b) The full tail, comprising two serially assembled BVSR segments. 
            (c) Vertebrae of various materials and diameters used in the experiments. 
            Experimental results showing the influence of the vertebra on (d) passive structural stiffness, (e) static output force, and (f) angular displacement under actuation.
        }
        \label{fig:3_TailDesign}
    \end{figure*}

    The BVSR tail is constructed with a hierarchical architecture. The fundamental building block is the \textit{BVSR joint}, whose geometric parameters are defined in Fig.~\ref{fig:3_TailDesign}(a) and listed in Table~\ref{tab:1_ParamBVSR}. A \textit{tail segment} is then formed by connecting three such joints in series. Finally, the complete \textit{4-DOF tail} consists of two independently actuated segments, as shown in Fig.~\ref{fig:1_Overview}(b).

    \begin{table}[htbp]
        \caption{Geometric Parameters of a Single BVSR Joint.}
        \label{tab:1_ParamBVSR}
        \centering
        \resizebox{0.45\columnwidth}{!}{%
        \begin{tabular}{llc}
            \hline
            \textbf{Symbol} & \textbf{Description} & \textbf{Value} \\ \hline
            \( H \)          & Height of SOA                     & 37.60 mm        \\
            \( H_{j} \)      & Height of single joint            & 42.51 mm        \\
            \( H_{a} \)      & Initial height between two panels & 36.51 mm        \\
            \( D_{a} \)      & Diameter of actuator circle       & 45.87 mm        \\
            \( D_{p} \)      & Diameter of end panels            & 70.94 mm        \\
            \( D_{b} \)      & Diameter of actuator base         & 24.32 mm        \\
            \( D_{c} \)      & Diameter of air channel           & 10.00 mm        \\
            \( D_{e} \)      & Diameter of vertebra              & 6.00 mm         \\
            \( \alpha_{1} \) & Angular location of actuator      & 72$^{\circ}$    \\
            \( \alpha_{2} \) & Tilt angle of segment 1           & 91.23$^{\circ}$ \\ \hline
        \end{tabular}%
        }
    \end{table}
    
    The actuator layout within each joint is functionally biomimetic, featuring a mirror symmetry across the sagittal (YZ) plane but an asymmetry across the transverse (XZ) plane. This arrangement prioritizes the motions most critical for inertial stabilization: flexion, extension, and waggle. To achieve a compact profile, the five actuators in each joint are arranged in a regular pentagon.
    
    The overall tail features a tapered, cone-like profile, with the diameter of the rigid panels decreasing from the root to the tip. The tail’s lightweight design benefits it in the horizontal mounting position by reducing gravitational droop and enabling low-inertia maneuvering. This complete assembly can be integrated onto various mobile platforms, such as the quadruped robot shown in Fig.~\ref{fig:1_Overview}(d), to enhance their dynamic capabilities.

\subsection{Performance Characterization of the Vertebra}

    A key hypothesis of this work is that the vertebra's mechanical properties directly govern the tail's performance. To validate this and guide our design, we conducted a parametric study comparing a segment with no vertebrae (``invertebrate'') to segments equipped with vertebrae of varying materials and diameters. We selected rods made of Rubber, TPU, Nylon, and PVC, with diameters of 4, 5, and 6 mm (Fig.~\ref{fig:3_TailDesign}(c)). The following tests were performed to quantify the vertebra's influence on stiffness, force, and motion.

\subsubsection{Structural Stiffness}

    To measure passive stiffness, the distal panel of the segment was displaced laterally using a linear motor stage (KLD FSL40-300), and the restoring force was measured with a force sensor (ARIZON 3031). As shown in Fig.~\ref{fig:3_TailDesign}(d), the invertebrate segment exhibits the most considerable deformation for a given force, confirming that the vertebra significantly reinforces the segment's lateral stiffness. Stiffness increases with both the Young's modulus of the vertebra's material and its diameter.

\subsubsection{Output Force}

    To measure static output force, the segment was actuated while its distal panel was connected to the force sensor. Figure~\ref{fig:3_TailDesign}(e) plots the output force against the differential input pressure ($\Delta$P). The vertebra provides essential structural support, preventing buckling and enabling the actuators to generate higher forces. The invertebrate segment produces the lowest force. Segments with stiffer vertebrae (Nylon, PVC) generate higher, more linear forces, indicating that a sufficiently stiff vertebra is crucial for stable and predictable force generation.

\subsubsection{Angular Displacement}

    To measure the motion workspace, the segment was actuated, and an IMU recorded its angular displacement. The results in Fig.~\ref{fig:3_TailDesign}(f) reveal a critical trade-off. The highly rigid vertebrae (PVC, 6 mm; Nylon, 6 mm) excessively constrain motion. Conversely, the invertebrate and softest-vertebra (Rubber, 6 mm) cases lack sufficient constraint, resulting in less predictable motion. The vertebrae with intermediate stiffness (e.g., Nylon, 5 mm; TPU, 6 mm) provide an optimal balance, guiding the segment's deformation to produce large and linear angular displacements.

    Based on this characterization, which demonstrates that the vertebra is a tunable design element for customizing performance, we selected the Nylon 6 mm rod for our final prototype. It offers the best compromise of structural reinforcement, linear force output, and an extensive, controllable range of motion.
    
\subsection{Fabrication and Assembly}

\begin{figure*}[htbp]
        \centering
        \includegraphics[width=1\linewidth]{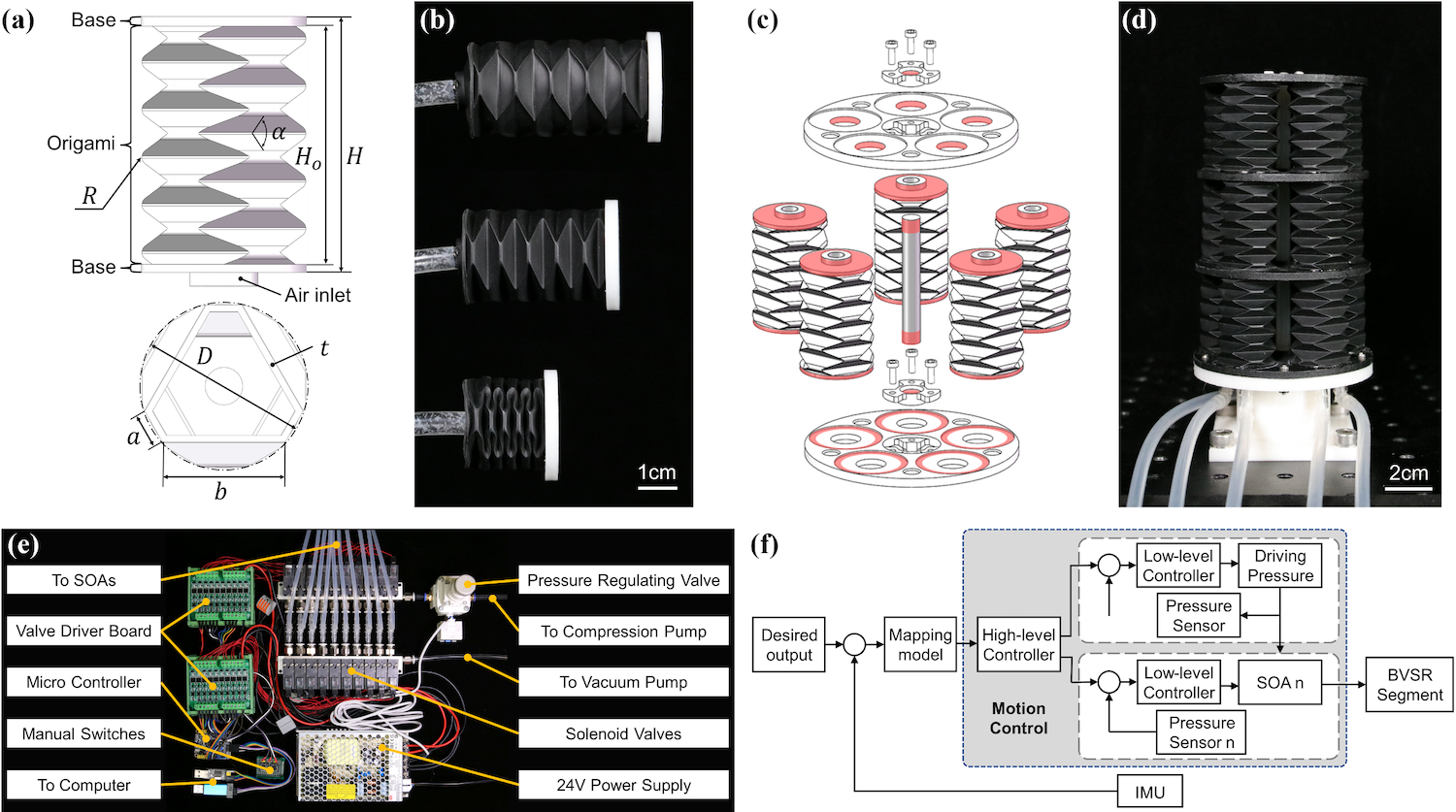}
        \caption{
            \textbf{Fabrication and actuation of the BVSR tail.}
            (a) Geometric parameters of the soft origami actuator (SOA). 
            (b) The actuator's linear elongation and contraction response to pressurization. 
            (c) The assembly process for a single BVSR joint. 
            (d) The fully assembled prototype of one BVSR segment. 
            (e) The pneumatic actuation system hardware. 
            (f) The cascaded motion and pressure control scheme.
        }
        \label{fig:4_TailActuation}
    \end{figure*}

    The prototype was realized using standard fabrication techniques. The rigid components (e.g., panels, connectors) were 3D-printed using MultiJet Fusion with Nylon PA12. The Soft Origami Actuators (SOAs) were fabricated using vacuum casting with polyurethane rubber. They feature an origami-inspired pattern on their walls \cite{liuotariidae, liu2024small} (Fig.~\ref{fig:4_TailActuation}(a)), which facilitates linear extension and contraction under positive and negative pressure, respectively (Fig.~\ref{fig:4_TailActuation}(b)). The geometric parameters of the SOA are listed in Table~\ref{tab:2_ParamSOA}. The motion of the SOA is shown in Supplementary Movie S1.

    \begin{table}[htbp]
        \caption{Geometric Parameters of the Soft Origami Actuator (SOA).}
        \label{tab:2_ParamSOA}
        \centering
        \resizebox{0.45\columnwidth}{!}{%
        \begin{tabular}{llc}
            \hline
            \textbf{Symbol} & \textbf{Description} & \textbf{Value} \\ \hline
            \( H \)          & Height of SOA                          & 37.60 mm        \\
            \( H_{o} \)      & Height of origami section              & 35.00 mm        \\
            \( t \)          & Thickness of origami facet             & 1.00 mm         \\
            \( \alpha \)     & Dihedral angle of trapezoid facets     & 73.83$^{\circ}$ \\
            \( R \)          & Fillet radius of short edge            & 1.20 mm         \\
            \( D \)          & Diameter of circumcircle               & 24.78 mm        \\
            \( a \)          & Length of short parallel side of facet & 5.30 mm         \\
            \( b \)          & Length of long parallel side of facet  & 18.21 mm        \\ \hline
        \end{tabular}%
        }
    \end{table}

    The assembly process is shown in Fig.~\ref{fig:4_TailActuation}(c). The SOAs were bonded to their dedicated bases on the rigid panels using a polyurethane (PU)-based adhesive. The selected vertebra was screwed into the center of the panels, creating a modular design that allows for easy replacement. The final assembled prototype of a single BVSR segment weighs 111.2 g in Fig.~\ref{fig:4_TailActuation}(d).

\subsection{Actuation and Control System}

\subsubsection{High-Pressure Pulsed Actuation Strategy}
    
    A primary limitation of conventional soft pneumatic actuators is their susceptibility to material failure at pressures exceeding 1.5-2 bar. Our system overcomes this by leveraging the vertebra as the primary load-bearing element against axial tensile forces. This allows the use of high-pressure pneumatic pulses (up to 6 bar) of short duration ($t < 100$ ms). This strategy maximizes the rate of energy input, generating high accelerations, while the limited duration prevents viscoelastic creep and catastrophic failure, enabling a performance regime previously inaccessible to untethered soft actuators. The negative pressure is limited to -95 kPa by the vacuum pump.

\subsubsection{Hardware and Control Architecture}

    The actuation hardware is shown in Fig.~\ref{fig:4_TailActuation}(e). It consists of an air compressor and a vacuum pump that feed a bank of twenty high-flow solenoid valves (OST Solenoid SY2/2N.C). These valves regulate air flow to the SOAs and are controlled by a microcontroller (STM32) via optical coupler equipment.

    The control architecture in Fig.~\ref{fig:4_TailActuation}(f) employs a cascaded, two-layer scheme \cite{liu2021a}. The inner, low-level loop performs pressure control for each SOA using feedback from dedicated pressure sensors. The outer, high-level loop controls the motion of the entire tail segment, using input from an IMU at the distal end to regulate its pose along the desired trajectory.

\section{Experiment Results}
\label{sec:Experiments}

\subsection{Performance Characterization}
\subsubsection{Characteristics of SOA}

\begin{figure}[htbp]
        \centering
        \includegraphics[width=1\linewidth]{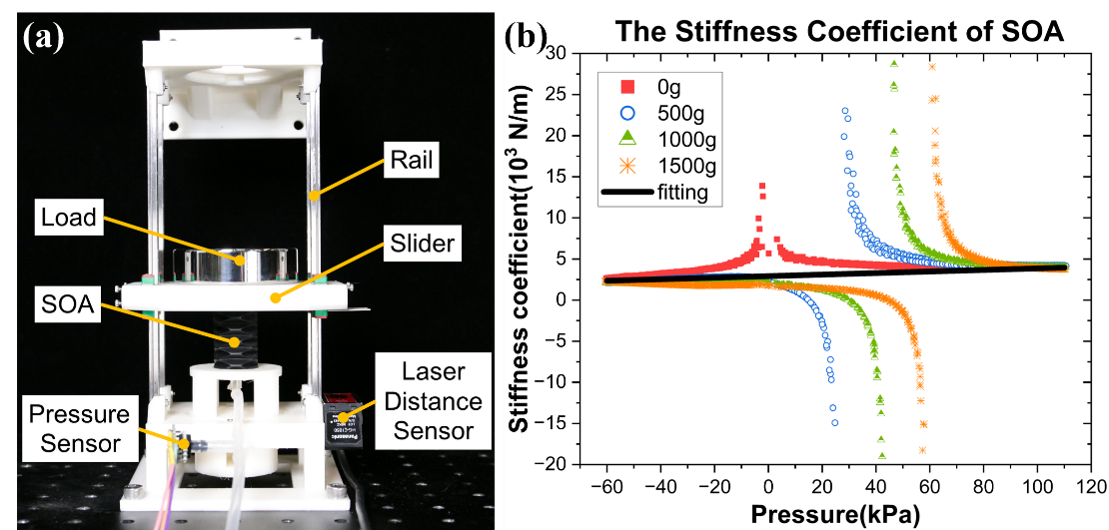}
        \caption{
        \textbf{Characteristics of SOA.}
        (a) The test setup for the stiffness coefficient of SOA. 
        (b) The experimental results were plotted and fitted to obtain the stiffness coefficient.}
        \label{fig:5_SOA}
    \end{figure}

    To obtain the stiffness coefficient \( k \) of the SOA, a series of tests were carried out on the actuator by adding different loads on one end, as depicted in Fig.~\ref{fig:5_SOA}(a). The loads were applied to the slider that compresses the SOA, including 0 g, 500 g, 1,000 g, and 1,500 g. When the input pressure balances the loads and the SOA keeps its original length, i.e., \( \Delta l_i \) is close to 0, the stiffness becomes very large according to Eq. (4), i.e., \( k = \frac{P_i A - F_i}{\Delta l_i} \). Therefore, the data around \( \Delta l_i = 0 \) were ignored when fitting the experimental results.  The tests were repeated three times for each load condition, with the displacement and pressure captured by designated sensors. A CMOS-type Micro Laser Distance Sensor (Panasonic HGC1050, 50 mm, 30 \textmu{}m resolution) was used to record the linear displacement of the slider. The results of the experiment are plotted in Fig.~\ref{fig:5_SOA}(b). According to the fitted curve, the stiffness coefficient is obtained as \( k = 9.4P + 2918 \), which is relative to the input pressure \( P \).

\subsubsection{Quasi-Static Workspace and Force Output}

\begin{figure}[t]
        \centering
        \includegraphics[width=1\linewidth]{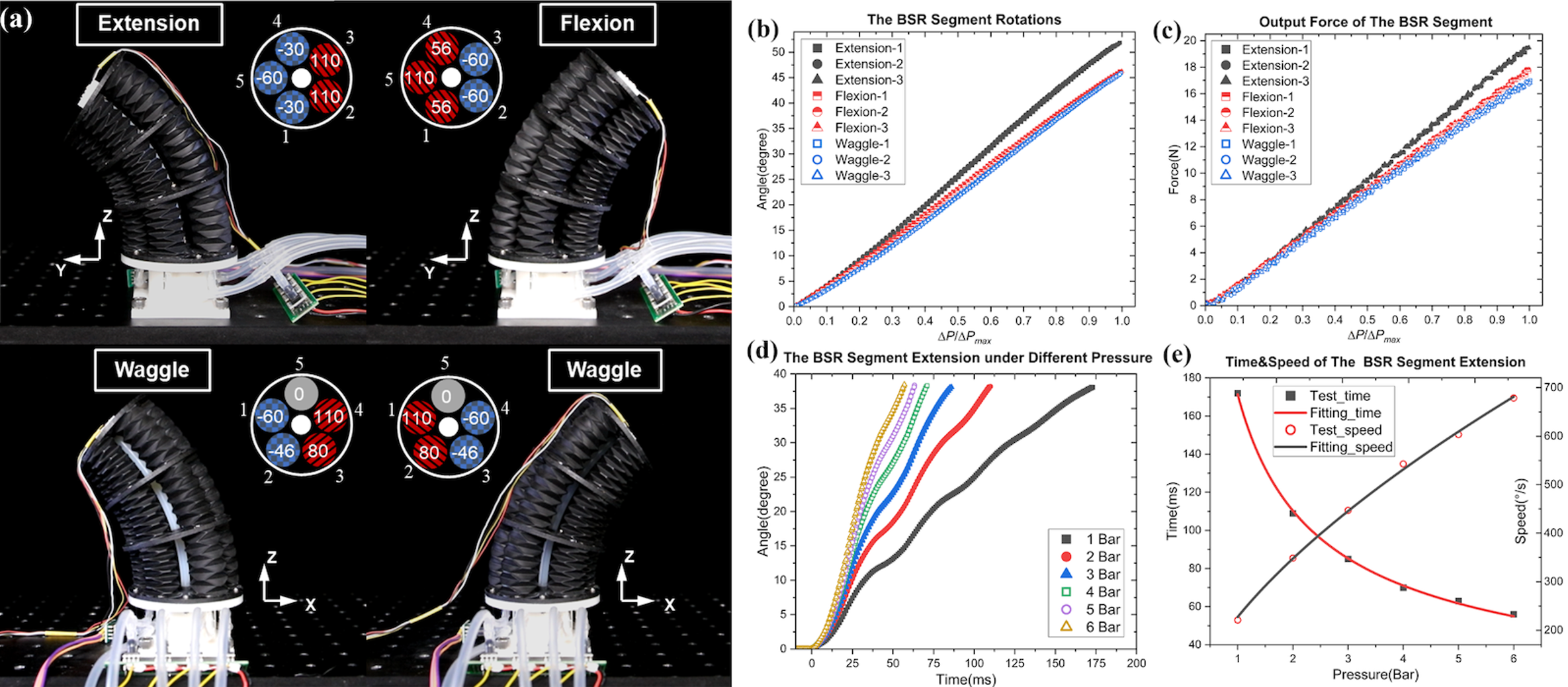}
        \caption{
            \textbf{Quasi-static and dynamic performance of the BVSR tail.}
            (a) The primary motions of a segment: extension, flexion, and waggle.
            (b) Quasi-static angular displacement vs. normalized pressure.
            (c) Quasi-static blocked force vs. normalized pressure.
            (d) Time-series data for a 38° extension motion under various driving pressures.
            (e) Empirical models relating motion time and velocity to driving pressure.
        }
        \label{fig:6_TailMovement}
    \end{figure}

    We first characterized the quasi-static performance of a single BVSR segment. The primary motions, such as extension, flexion, and waggle, are shown in Fig.~\ref{fig:6_TailMovement}(a). The angular workspace and maximum blocked force were measured for each motion, with results plotted in Figs.~\ref{fig:6_TailMovement}(b-c) and summarized in Table~\ref{tab:3_QSOutput}. The motions of the segment are shown in Supplementary Movie S1. The segment achieves a workspace of over 46$^{\circ}$ in all directions and generates forces up to 19.7 N. Notably, both the range of motion and force output are greatest in the extension direction. This anisotropic performance, a direct result of the asymmetric actuator arrangement, is advantageous for counteracting gravity when the tail is mounted horizontally.

    \begin{table}[htbp]
        \centering
        \caption{Quasi-Static Performance of a Single BVSR Segment.}
        \label{tab:3_QSOutput}
        \begin{tabular}{lcc}
            \hline
            \textbf{Motion Direction} & \textbf{Max Angle} & \textbf{Max Force} \\ \hline
            Extension                 & 51.9$^{\circ}$     & 19.7 N             \\
            Flexion                   & 46.5$^{\circ}$     & 17.8 N             \\
            Waggle                    & 46.1$^{\circ}$     & 16.8 N             \\ \hline
        \end{tabular}
    \end{table}

\subsubsection{Dynamic Performance Enhancement}

    A key feature of the BVSR tail is its ability to operate under high pressure, enabling high-speed motion. We tested the segment's velocity using pressures from 1 to 6 bar in the setup shown in Fig.~\ref{fig:6_TailMovement}(a). During the tests, the tail was actuated under the specified pressure and allowed to move freely without any manual fixation. The angular velocity was recorded during the motion before the bending angle passed through 38°. For example, at an actuation pressure of 6 bar, the tail can eventually reach a much larger bending angle, but only the motion segment up to 38° was selected for velocity evaluation to provide a consistent basis for comparison across different pressure conditions.  As shown in Fig.~\ref{fig:6_TailMovement}(d), higher pressures drastically reduce the time required to complete the motion. At 6 bar, the segment achieves peak angular velocities of 678.6$^{\circ}$/s (extension), 628.1$^{\circ}$/s (flexion), and 625.0$^{\circ}$/s (waggle), representing velocity increases of over 200$\%$ compared to the 1 bar case. The relationship between input pressure \(P\) and both motion time \(T\) and velocity \(V\) was empirically modeled (Fig.~\ref{fig:6_TailMovement}(e)), yielding power-law relationships (\(T \propto P^{-0.63}\), \(V \propto P^{0.61}\)) that are crucial for predictive control.

\subsubsection{Inertial Force and Torque Generation}

\begin{figure}[t]
        \centering
        \includegraphics[width=1\linewidth]{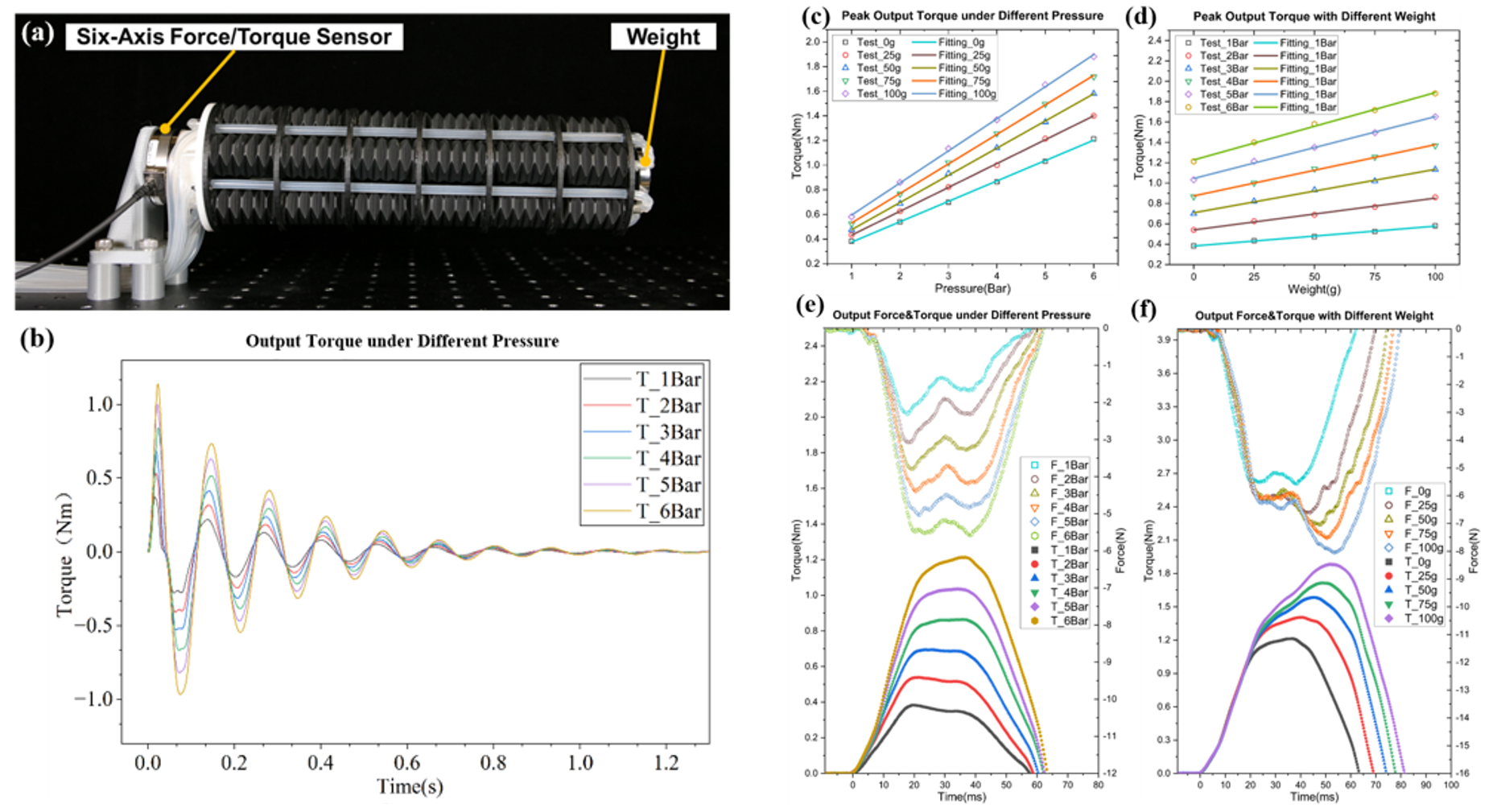}
        \caption{
            \textbf{Inertial output performance of the BVSR tail.}
            (a) Experimental setup for measuring the full tail's inertial output.
            (b) Inertial force and torque output as a function of driving pressure.
            (c) Inertial output as a function of added distal mass.
            (d) The linear relationships between peak torque and driving pressure and distal load, respectively.
            (e) Output inertial forces and moments of the tail with different distal weights were recorded and plotted.
        }
        \label{fig:7_InertialOutput}
    \end{figure}

    The primary function of the tail is to generate inertial forces and torques for platform stabilization. We measured these outputs using a full two-segment tail mounted on a six-axis force sensor (ATI Mini45, 0-145 N, 0-5 Nm) as shown in Fig.~\ref{fig:7_InertialOutput}(a). The representative raw data of output torque under different pressures is presented as shown in Fig..~\ref{fig:7_InertialOutput}(b). The curves show attenuation oscillations during tail-swing motions, generated by the pulse-like pneumatic input. The results shown in Fig.~\ref{fig:7_InertialOutput}(c-f) confirm that the inertial output scales directly with both actuation pressure and distal mass. Increasing the driving pressure from 1 to 6 bar boosted the peak output force by 139.9$\%$ (to 5.58 N) and the peak torque by 216.7$\%$ (to 1.21 Nm), as shown in Fig.~\ref{fig:7_InertialOutput}(e). Furthermore, adding a 100 g mass to the tip at 6 bar pressure increased the peak torque to 1.88 Nm and the peak output force to 8.06 N, generating 55.1$\%$ and 44.4$\%$ increases, respectively, over the non-loaded case (Fig.~\ref{fig:7_InertialOutput}(f)). The linear relationships between peak torque and both pressure and mass (Fig.~\ref{fig:7_InertialOutput}(c-f)) demonstrate that the tail's inertial effects are highly tunable and controllable. 

\subsubsection{Customizable Performance}

\begin{figure}[htbp]
        \centering
        \includegraphics[width=1\linewidth]{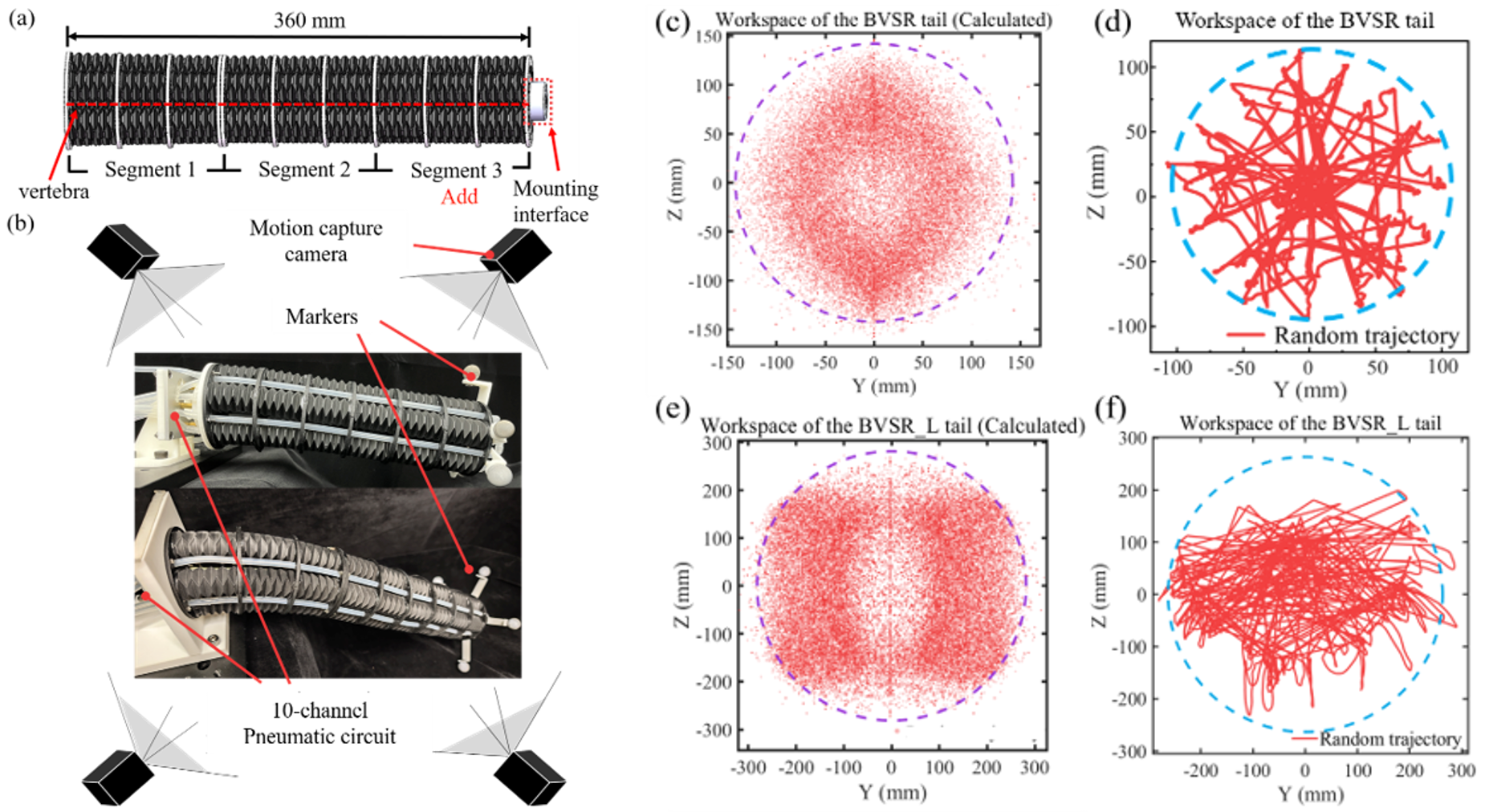}
        \caption{
            \textbf{Workspace comparison of the BVSR tail.}
            (a) Design of the extended BVSR\_L tail.
            (b) Experimental setup used to measure the tails' workspaces with motion capture equipment.
            (c) Calculated workspace of the BVSR tail.
            (d) Measured workspace of the BVSR tail.
            (e) Calculated workspace of the BVSR\_L tail.
            (f) Measured workspace of the BVSR\_L tail, obtained from the circumscribed circle of random movement recordings.          
        }
        \label{fig:8_Workspace}
    \end{figure}

    Apart from the vertebra's design and material, the BVSR tail's workspace is also related to its diameter-to-length ratio. To demonstrate the workspace's customization capability, we designed and fabricated a new tail, BVSR\_L. The design of the BVSR\_L is presented in Fig.~\ref{fig:8_Workspace}(a). An additional BVSR segment was integrated at the distal end of the tail. The added segment adopts the same modular architecture as the original design, consisting of three serially connected BVSR joints with identical rigid panels, SOAs, and a vertebra. Consequently, the BVSR\_L tail comprises three BVSR segments (nine BVSR joints in total), yielding an overall length of approximately 360 mm while preserving the original actuation principle and structural configuration. 
    
    We calculated the workspaces of the two tails using the kinematic model, with the results plotted in Fig.~\ref{fig:8_Workspace}(c) and (e). We carried out experiments on the BVSR and BVSR\_L tails to record workspaces. As shown in Fig.~\ref{fig:8_Workspace}(b), the tails are fixed horizontally; a set of motion capture equipment (Mars 1.3H, NOKOV) records the trajectory of the tails' distal end. For the BVSR, the pressure in each chamber is randomly assigned within the following range: 101 kPa to 150 kPa in the pressurized state, 50 kPa to 101 kPa in the depressurized state, and 101 kPa in the inactive state. The tail performs this process periodically, with each cycle spaced 1 second apart and repeated 60 times. The system completed pressure and trajectory data collection at 120 Hz, recording 18 sets of data, totaling 129,600 pairs (60\(\times\)120\(\times\)18). For the BVSR\_L, the pressure in each chamber is randomly allocated within the following range: 101 kPa to 251 kPa in the pressurized state, 21 kPa to 101 kPa in the depressurized state, and 101 kPa in the inactive state. The tail performs this process periodically, with each cycle spaced 1 second apart and repeated 60 times. The system collected pressure and trajectory data at 100 Hz, recording 20 sets totaling 120,000 pairs. The experimental data of BVSR and BVSR\_L are plotted in Fig.~\ref{fig:8_Workspace}(d) and (f), respectively. 
    
    According to the data, the BVSR's workspace is about \(\pm\) 100 mm, while the BVSR\_L tail's workspace is about \(\pm\) 300 mm, which agrees with the corresponding calculated workspaces. The tail's inertial output can be customized by the tip mass, as shown in Fig.~\ref{fig:7_InertialOutput}, which can be further utilized in the tail installed on the robotic platform. Building on this, if a specific range of motion and inertial output are required, we can customize the tail to meet those requirements.  
   
\subsection{Validation of Analytical Models}

    We performed a series of experiments to validate the analytical models developed in Section~\ref{sec:Concept}. A comparative summary of the BVSR tail's performance against other state-of-the-art robotic tails is provided in Table~\ref{tab:4_TailCompare}.

    \begin{table*}[htbp]
        \centering
        \caption{Comparison with State-of-the-Art Robotic Tails.}
        \label{tab:4_TailCompare}
        \resizebox{\linewidth}{!}{%
            \begin{tabular}{lcccccccc}
                \hline
                \textbf{Robotic Tail} & \textbf{Weight (g)} & \textbf{Length (cm)} & \textbf{Rigidity} & \textbf{Actuation} & \textbf{DOF} & \textbf{Speed ($^{\circ}$/s)} & \textbf{Torque (Nm)} & \textbf{Angle ($^{\circ}$)} \\ \hline
                \textbf{BVSR tail (This work)} & \textbf{283} & \textbf{21} & \textbf{Soft} & \textbf{Pneumatic} & \textbf{4} & \textbf{678.6} & \textbf{1.2; 2.48 (payload 500 g)} & \textbf{\(>\)45} \\
                Active tail \cite{heim2016on} & 31 & 16.8 & Rigid & Motor & 1 & - & - & - \\
                Kangaroo tail \cite{santiago2016soft} & - & - & Rigid \& Flexible & Motor & 1 & - & 2.01 & 92 \\
                Lizard tail \cite{changsiu2011a} & 17 & 12.7 & Rigid & Motor & 1 & - & 0.005 & - \\
                Continuum tail \cite{rone2014continuum} & 2250 & 50 & Flexible & Motor & 2 & - & 22.6 & - \\
                Cable-driven soft tail \cite{persons2011the} & 200 & 40 & Rigid \& Flexible & Motor & \(>\)2 & - & - & - \\
                Lightweight tail \cite{butt2021modeling} & 110 & 100 & Rigid \& Flexible & Motor & 1 & 64.2 & 2 & 90 \\ 
                Prehensile tail \cite{XM2025softtail} & 60 & 21 & Soft & Pneumatic\& Motor & 3 & 428.6 & 0.45 & 180 \\ 
                R3RT \cite{saab2019design} & 900 & 500 & Flexible & Motor & \(>\)3 & - & - & 270 \\ 
                USRT \cite{rone2018design} & 510 & 480 & Flexible & Motor & 6 & 180 \(/\)225 & - & 245 \\ \hline
            \end{tabular}%
        }
    \end{table*}

\begin{figure}[h]
        \centering
        \includegraphics[width=0.7\linewidth]{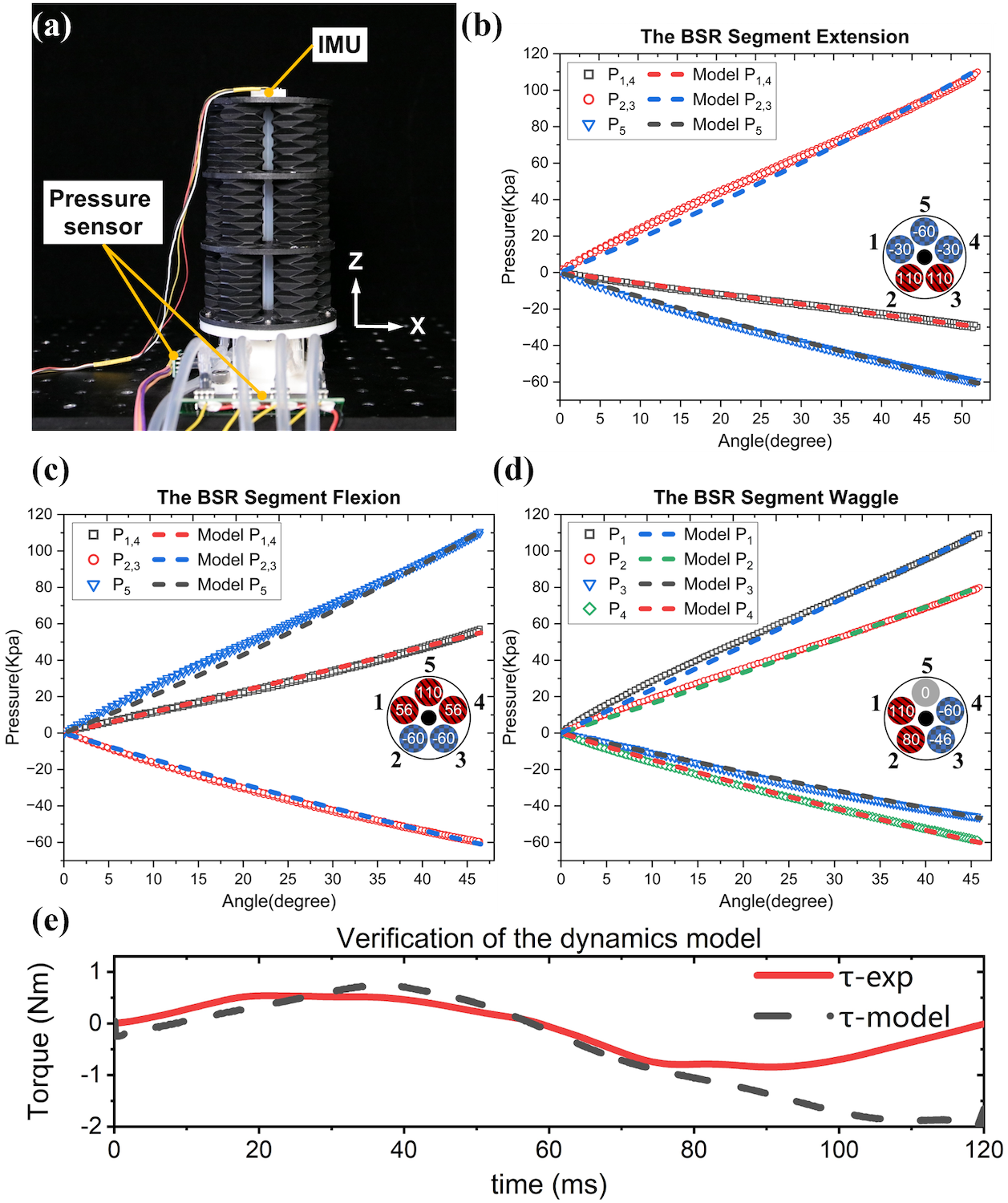}
        \caption{
            \textbf{Verification of the analytical models.}
            (a) Experimental setup for model validation.
            (b-d) Comparison of experimental pressure data (points) with predictions from the inverse static model (Eq.~\eqref{eq:5}, lines) for flexion, extension, and waggle motions.
            (e) Comparison of measured inertial torque (experimental) with the forward dynamic model's prediction (Eq.~\eqref{eq:6}).
        }
        \label{fig:9_VerifyKinematics}
    \end{figure}

    \textit{First}, to validate the inverse kinematic model presented in Eq.~\eqref{eq:5}, we measured the pressures and angular displacement in a series of quasi-static experiments and compared them to the model's predictions, with the experiment setup shown in Fig.~\ref{fig:9_VerifyKinematics}(a). The tail output extension, flexion, and waggle motions in respective experiments were plotted in Figs.~\ref{fig:9_VerifyKinematics}(b-d).  In each experiment, the input pressure was gradually increased in small increments from 0 kPa, and the actuator was allowed to reach steady-state deformation at each pressure level before data acquisition. The rotational angle of the BVSR segment in each rotation was recorded by the IMU module JY931 with MPU6050 (WitMotion, 0.1$^{\circ}$ accuracy) fixed in the distal center of the segment, and the pressure input for each SOA was recorded by the dedicated pneumatic pressure sensor SSCDANN060PAAA5 (Honeywell, 413 kPa). As shown in Figs.~\ref{fig:9_VerifyKinematics}(b-d), the analytical results show good agreement with the experimental data across all primary motions. Minor disparities at the onset of motion under high pressure can be attributed to the viscoelastic hysteresis of the soft actuators, a well-known effect not captured in the static model. The motions of the BVSR tail are shown in Supplementary Movie S1.

    \textit{Next}, we validated the dynamic model in Eq.~\eqref{eq:6} using data from a 2-bar flexion movement. The theoretical torque \(\tau_{model}\) was calculated using the model, with the damping coefficient \(c\) and a pressure profile parameter \(t_0\) optimized to minimize the RMSE between the predicted and measured torque. The maximum positive input pressure was 200 kPa, and the vacuum pressure was -90 kPa. Since the pressure gradually increased from 0 to the maximum \(P_0\), a polynomial function:
    \begin{equation} \label{eq:13}
        P(t) =P_{0}t^2(\frac{3}{t_{0}^{2}}-\frac{2}{t_{0}^{3}}t), \quad \text{where } 0<t<t_0,
    \end{equation} 
    was employed to represent the progression of the pressure increase from start to \(t_0\). The pressure function \(P(t)\) satisfies \(P(t_0)=P_0\) and \(P'(0)=P'(t_0)=0\) to ensure a smooth growth progression. The time \(t_0\) and the damping coefficient \(c\) were decided by minimizing the RMSE between the predicted and measured torque, i.e., \(t_0=40\) ms and \(c=52\). The resulting model captures the initial dynamic response and overall trend of the experimental data (RMSE = 0.6829), as shown in Fig.~\ref{fig:9_VerifyKinematics}(e). We believe the discrepancy observed after approximately 80 ms is primarily due to the viscoelastic hysteresis of the silicone-based SOAs, which is not explicitly accounted for in the model. The model describes the BVSR joint using a pressure-dependent elastic restoring force together with linear viscous damping. While this formulation captures the dominant dynamic behavior during the initial stage of motion, it inherently assumes an instantaneous elastic response and therefore does not account for the viscoelastic hysteresis of the silicone material. Since viscoelastic hysteresis is a time-dependent material property, its influence accumulates during the motion. Consequently, the model predicts the initial dynamic response with good accuracy, whereas after approximately 80 ms, the experimentally measured torque decays more slowly than the model prediction, leading to the increasing discrepancy observed in Fig.~\ref{fig:9_VerifyKinematics}(e). Therefore, this discrepancy primarily reflects the limitations of the current mechanical representation of the SOA rather than of the dynamic formulation. Incorporating a more sophisticated viscoelastic model will be considered in future work to further improve the prediction accuracy.
    \begin{figure}[htbp]
        \centering
        \includegraphics[width=1\linewidth]{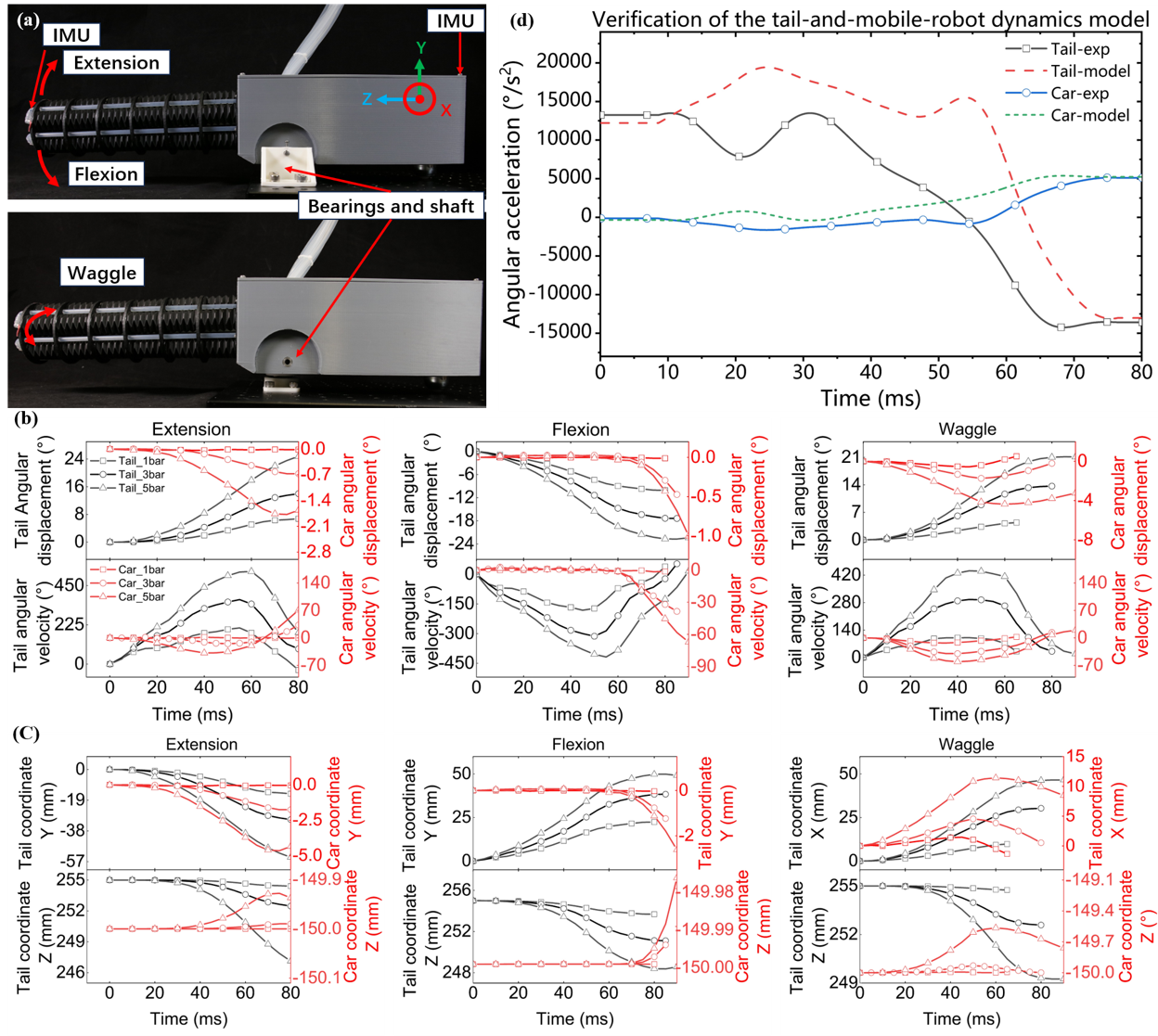}
        \caption{
            \textbf{Validation of the coupled tail-platform dynamics.}
            (a) Experimental setup with the tail mounted on a hinged cart.
            (b) Measured angular displacement and velocity of the cart (red) and tail (blue) under different driving pressures and motions.
            (c) Positions of the cart (red) and tail (blue). 
            (d) Representative comparison between the theoretical (line) and experimental (points) angular acceleration of the cart.
        }
        \label{fig:10_VerifyTail}
    \end{figure}
    
    \textit{Finally}, we verified the simplified tailed-platform model in Eqs.~\eqref{eq:21}\&\eqref{eq:22} using the setup in Fig.~\ref{fig:10_VerifyTail}(a). The rear wheels and chassis of a cart are hinged on rotating shafts, respectively, to investigate the effects of the BVSR tail's motion direction and amplitude on the platform's posture. The shafts are mounted on bearings to minimize friction. The BVSR tail was actuated by driving pressures of 1, 3, and 5 bar in extension, flexion, and waggle motion, respectively, and the angular motions of both the tail and the cart were recorded in Fig.~\ref{fig:10_VerifyTail}(b). For direct interpretation of the input and output, the positions of the tail and the cart are calculated from the angular displacement and plotted in Fig.~\ref{fig:10_VerifyTail}(c). As predicted by conservation of momentum, the cart rotates in the direction opposite to the tail's angular acceleration. The front wheel of the cart was lifted at the beginning of extension and the end of flexion. The cart turned at the beginning of the waggle. The amplitudes of the angular displacement and velocity increase with increasing driving pressure. The maximum angular displacements of the cart reach 1.8$^{\circ}$ during extension and 4.4$^{\circ}$ during waggle of the tail under 5 bar driving pressure. The extension, flexion, and waggle motions are all governed by the same dynamic model, with the only difference being the actuation pressure distribution.  As a representative case, a comparison between the measured cart acceleration (5-bar extension experiments) and the model's prediction is shown in Fig.~\ref{fig:10_VerifyTail}(d).  Although the model predictions match the experimental data reasonably well at the beginning and end of the motion, noticeable discrepancies are evident in the intermediate angular-acceleration trajectory. The model includes the dominant inertial, damping, and coupling effects, based on assumptions including idealized kinematics, constant model parameters, and simplified joint deformation. In practice, the dynamic response is also influenced by parameter variations, friction, transmission compliance, and higher-order deformations that are not explicitly captured in the analytical formulation. These factors can alter the system's effective inertial and dissipative characteristics, leading to deviations from the predicted angular-acceleration trajectory. As a result, the model can reproduce the overall motion and boundary conditions with reasonable accuracy, thereby validating its potential utility for predicting the tail's inertial effects on a mobile platform. 
    
\subsection{System-Level Functional Demonstrations}

\subsubsection{Dynamic Motion Repeatability}

\begin{figure}[htbp]
        \centering
        \includegraphics[width=0.5\linewidth]{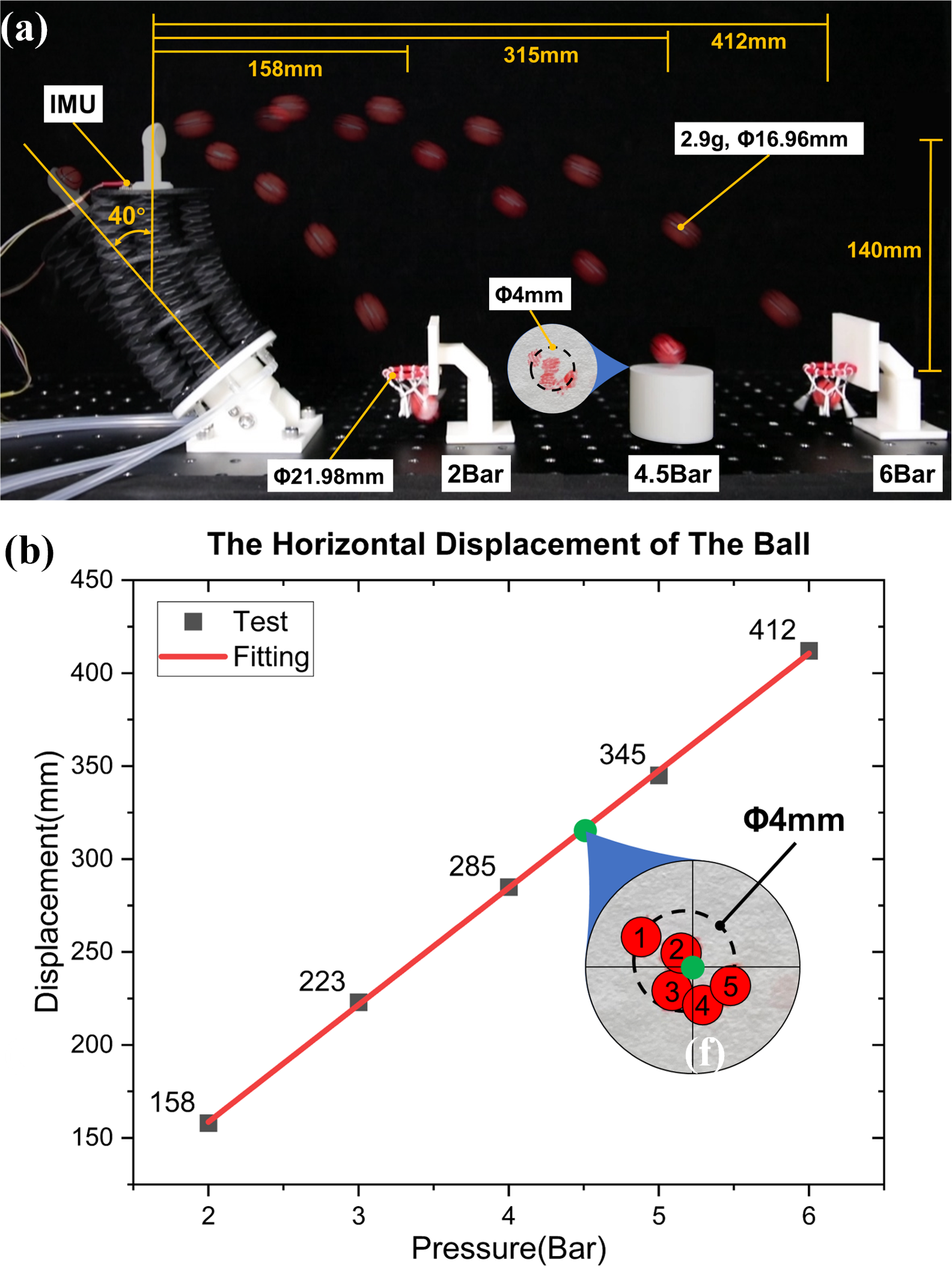}
        \caption{
            \textbf{Demonstration of dynamic precision.}
            (a) The ball shooting setup. The inset shows the tight grouping of consecutive shots.
            (b) The linear relationship between driving pressure and shot distance, highlighting the system's controllability.
        }
        \label{fig:11_VerifyShooting}
        \vspace{-0.4cm}
    \end{figure}

    A ball-shooting task was devised to test the tail's dynamic repeatability. A BVSR segment was tilted by 40$^{\circ}$, and a small ball (2.9 g) placed in the spoon at the distal end of the segment was shot into the basket by the extension motions (Fig.~\ref{fig:11_VerifyShooting}(a), Supplementary Movie S2). 
    The horizontal distances of the shot scaled linearly with input pressures as shown in Fig.~\ref{fig:11_VerifyShooting}(b). In 5 consecutive trials at 4.5 bar, all shots landed within a tight grouping, with a mean radial error of less than 4 mm, demonstrating the tail's capacity for repeatable and high-speed open-loop tasks.

\subsubsection{Mobile Platform Assistance}

\begin{figure}[htbp]
        \centering
        \includegraphics[width=0.7\linewidth]{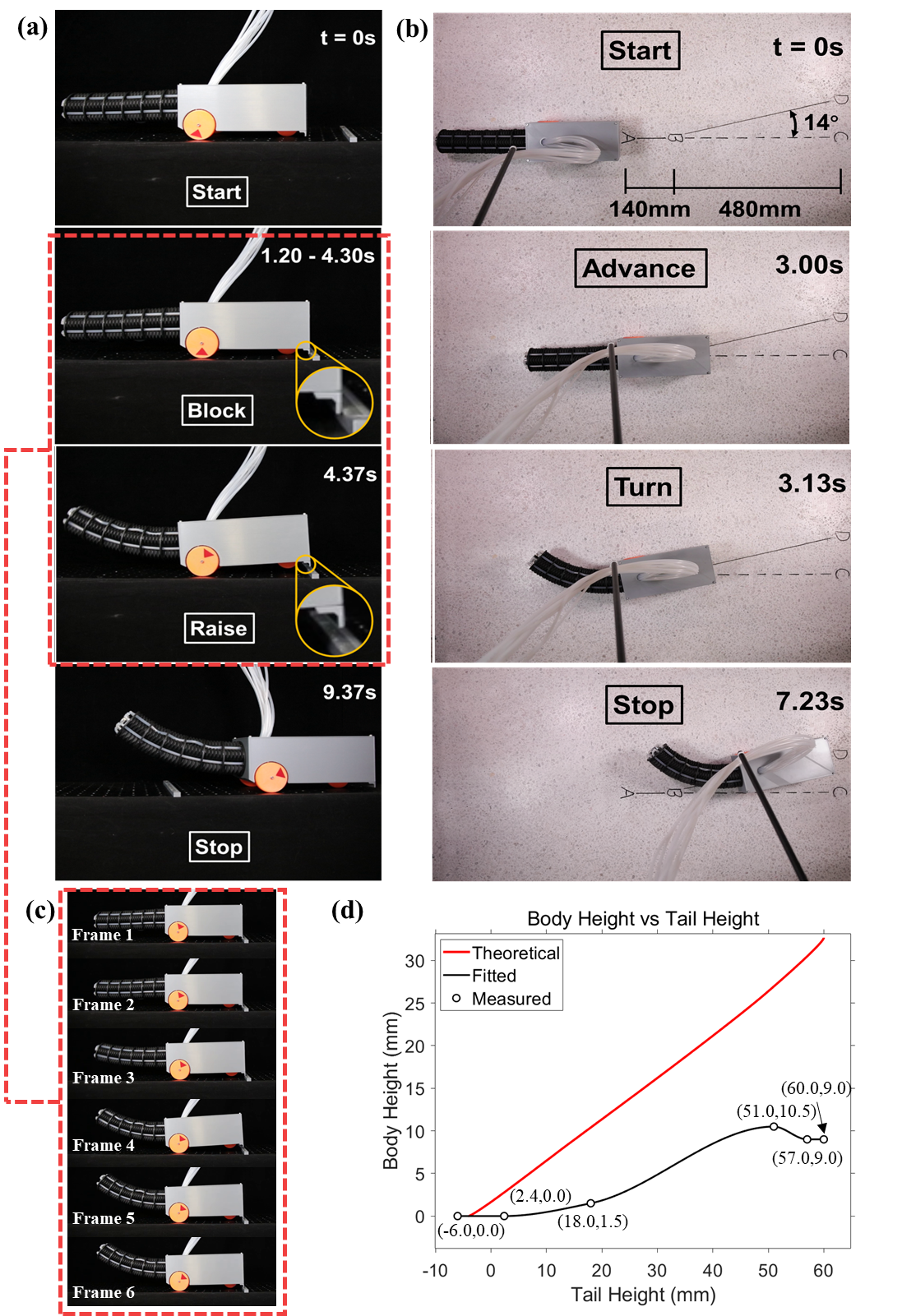}
        \caption{
            \textbf{Demonstrations of the BVSR tail providing inertial assistance to a mobile platform.}
            (a) Obstacle crossing via a vertical tail flick. 
            (b) Directional steering via a horizontal wag.           
            (c) 6 image frames from the video recording capture the tail and cart lifts. 
            (d) Relationship of the tail and cart lift height is measured and plotted to compare with the theoretical model.
        }
        \label{fig:12_TailCart}
    \end{figure}

    We mounted the tail on a wheeled cart (weight 700 g) to demonstrate its ability to provide meaningful inertial assistance. In an obstacle-crossing task (Fig.~\ref{fig:12_TailCart}(a), Supplementary Movie S3), 
    a rapid upward flick of the tail generated sufficient torque to lift the cart's front wheel over a 9 mm obstacle. In a steering task (Fig.~\ref{fig:12_TailCart}(b)), a horizontal wag of the tail induced a 14$^{\circ}$ change in the cart's heading. These tests confirm that the tail can generate functionally significant forces to manipulate a mobile platform's posture and trajectory. To provide quantitative validation, we measured the vertical lift distances of the tail's tip and the front end of the cart from the obstacle-crossing video recording. The time period of the vertical tail flick is very short; 6 image frames of the video are shown in Fig. ~\ref{fig:12_TailCart}(c), capturing the lifts of the tail and cart. By measuring the height of the tail's and cart's lift in each frame, the relationship is obtained and plotted in Fig.~\ref{fig:12_TailCart}(d). The curve shows that the cart is lifted by the tail's extension motion to a height of up to 10.5 mm, so that the obstacle is crossed over the top. The lifted height is calculated by the theoretical model in Eqs.~\eqref{eq:20}-\eqref{eq:22} and plotted in Fig.~\ref{fig:12_TailCart}(d). Compared with the experimental data, the model overestimated the cart's lift height, while the trends are similar. This indicates that BVSR tail compliance affects the transmission of the tail's dynamic input to the cart.

\subsubsection{Integration with a Quadruped Robot}

\begin{figure*}[htbp]
        \centering
        \includegraphics[width=1\linewidth]{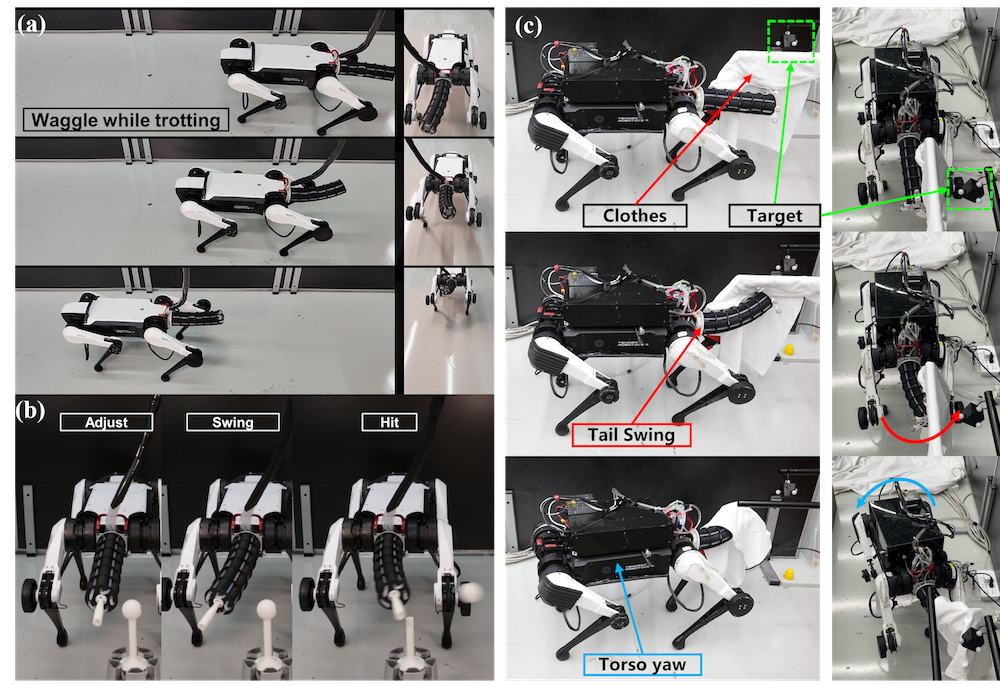}
        \caption{
            \textbf{The integration on a quadruped.}
            (a) The BVSR tail integrated onto a quadruped robot, demonstrating independent motions during trotting, 
            (b) collaborative manipulation to hit a ball, 
            (c) coordinated motions of the BVSR tail on the quadruped to pick up the soft clothes in an unstructured environment.
        }
        \label{fig:13_TailQuadTailCloth}
    \end{figure*}

    We integrated the tail onto a quadruped robot (MAX, Tencent Robotics X, 14 kg) to demonstrate its compatibility and potential for collaborative tasks in Fig.~\ref{fig:13_TailQuadTailCloth}. 
    The tail moved independently during the robot's trotting gait (Supplementary Movie S4) and worked in conjunction with the robot to position itself for a strike. Furthermore, it assisted the robot's whole-body motion to pick up a piece of cloth from the ground (Supplementary Movie S5). These demonstrations validate the BVSR tail as a practical, versatile, and high-performance appendage for advanced robotic systems. 

\subsubsection{Human-tail Interaction}

\begin{figure}[htbp]
        \centering
        \includegraphics[width=0.5\linewidth]{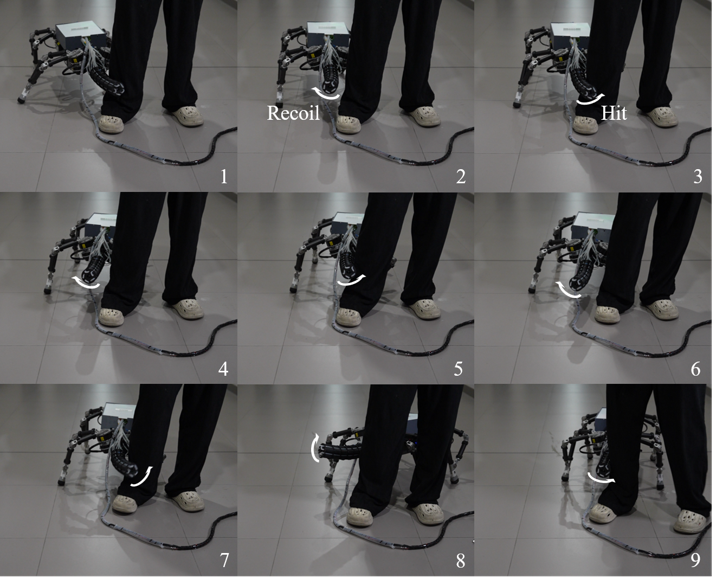}
        \caption{
            \textbf{Human-tail direct interaction.}
            The BVSR tail, installed on the quadruped, repeatedly contacts a person's leg, indicating preliminary compliant contact behavior.
        }
        \label{fig:14_HumanTailDirect}
    \end{figure}

    To evaluate the tail's compliant interaction with humans and the environment, a new quadruped (asOverdog, AncoraSpring, 14.5 kg) was used to install the BVSR tail, and two interactive experiments were conducted. In the human-tail direct interaction experiment, as shown in Fig.~\ref{fig:14_HumanTailDirect}, a person stood within the workspace of the BVSR tail. The tail wagged in coordination with the quadruped robot, making repeated contact with the person's leg (Supplementary Movie S6). Throughout the process, the BVSR tail remained fully functional and conformed its shape in response to the physical contacts. The direct interactions caused no discomfort or injury to the person, providing preliminary evidence of compliant contact behavior in this direct human-robot interaction. 

\begin{figure}[htbp]
        \centering
        \includegraphics[width=0.5\linewidth]{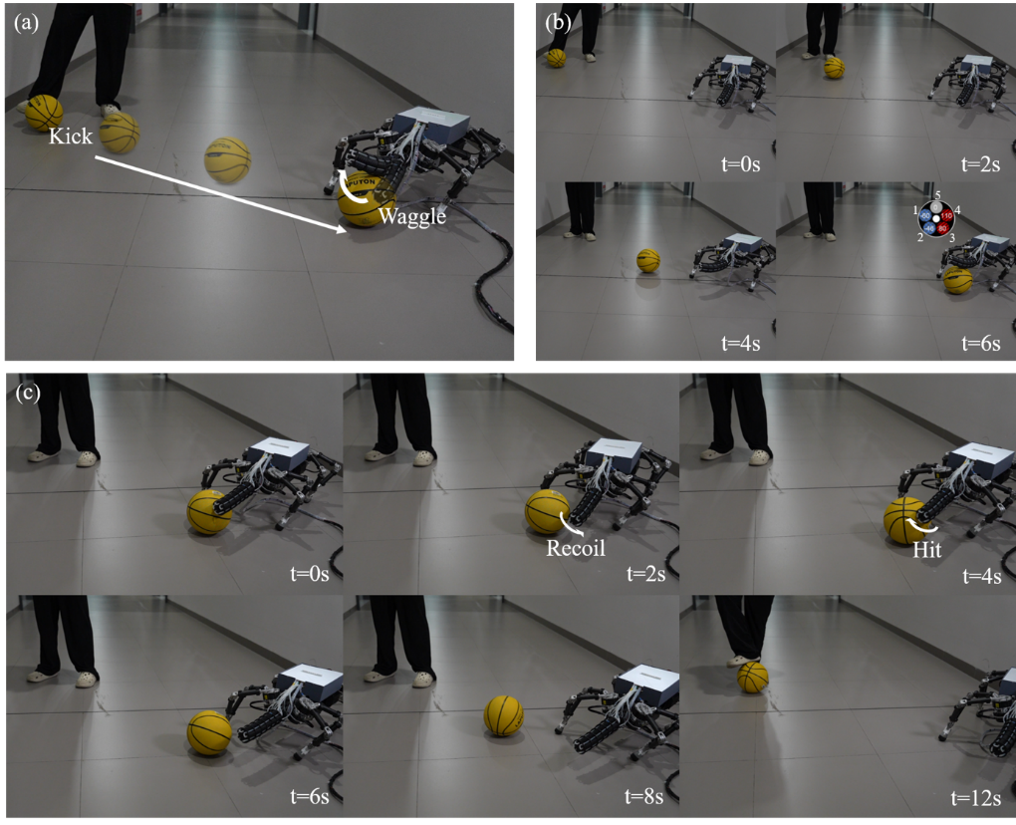}
        \caption{
            \textbf{Human-tail indirect interaction.}
            (a) The person kicks the ball to the tailed-quadruped, and the quadruped stops the ball with the BVSR tail (b). (c) The quadruped kicks the ball back to the person using the tail.
        }
        \label{fig:15_HumanTailIndirect}
    \end{figure}

    The human-tail indirect interaction was carried out as shown in Fig.~\ref{fig:15_HumanTailIndirect}. In this experiment, the person kicked a basketball toward the tail, which slightly lifted and actively curved into an arc to stop the ball. Subsequently, the quadruped robot swung the tail to return the ball to the person (Supplementary Movie S7). This indirect interaction with a human demonstrates the tail's functional potential for use with everyday objects due to its compliance. The change of platform also shows the tail's applicability across different robotic platforms. 

\subsubsection{Impact Resistance}

\begin{figure}[htbp]
        \centering
        \includegraphics[width=1\linewidth]{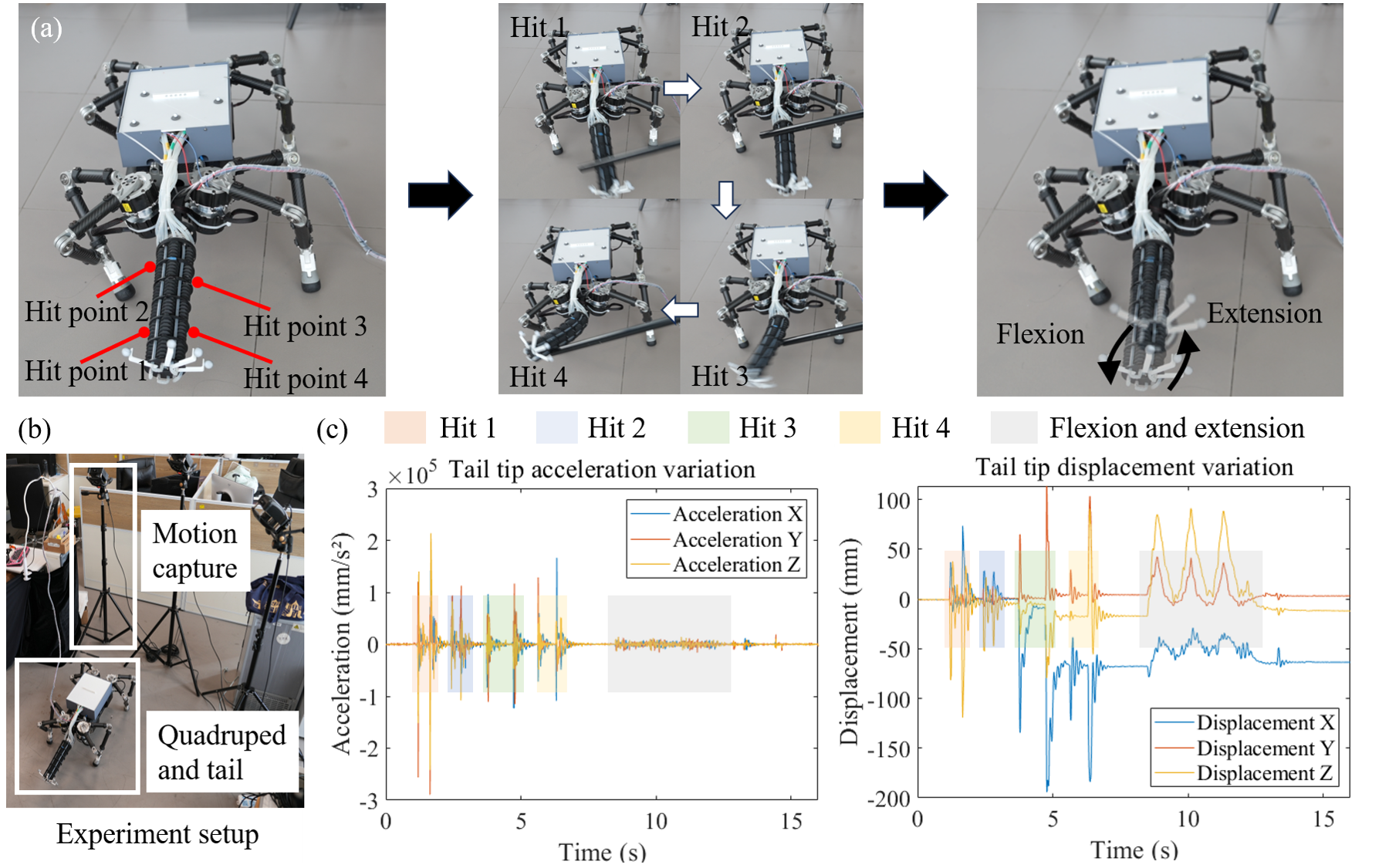}
        \caption{
            \textbf{Impact resistance test of the BVSR tail.}
            (a) Experimental setup, including the impact locations and testing procedure. 
            (b) Experimental results, including the acceleration and displacement responses of the tail tip, and the post-impact flexion and extension motions.
        }
        \label{fig:16_ImpactResist}
    \end{figure}

    To evaluate the impact robustness and functional resilience of the BVSR, an impact resistance test was conducted as illustrated in Fig.~\ref{fig:16_ImpactResist}(a). Four impact locations (Hit points 1–4) were selected along the tail, and a rigid rod was used to strike each location sequentially (Supplementary Movie S8). Each impact point was struck twice. After the impact tests, the tail was actuated to perform both flexion and extension motions to assess whether its motion capability was affected. Meanwhile, IMUs mounted on the tail tip and the back of the quadruped, and a set of motion capture equipment (Mars 1.3H, NOKOV) as shown in Fig.~\ref{fig:16_ImpactResist}(b), were used to record the acceleration and displacement responses during the tests. 
    
    The experimental results are presented in Fig.~\ref{fig:16_ImpactResist}(c). During the impacts, the peak acceleration measured at the tail tip reached approximately 200 m/s². Despite repeated impacts, the BVSR remained fully functional and executed both flexion and extension motions without observable structural damage or performance degradation. Furthermore, only minor displacement and posture variations were observed in the quadruped body, indicating that external impacts on the tail had negligible influence on the robot's stability. These results demonstrate the excellent compliance, impact tolerance, and resilience of the proposed BVSR. 

\subsubsection{Disturbance Absorption}

\begin{figure}[htbp]
        \centering
        \includegraphics[width=1\linewidth]{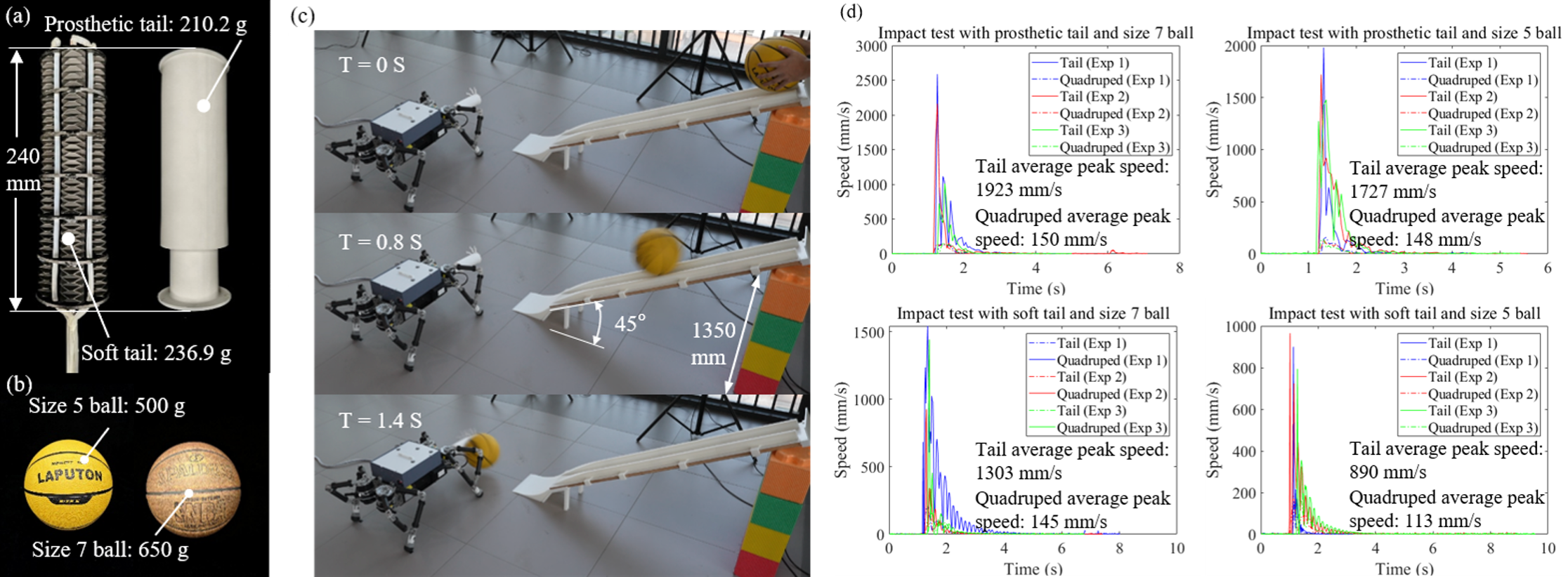}
        \caption{
            \textbf{Effect of tail compliance on the quadruped posture stability.}
             (a) A prosthetic tail is fabricated to compare with the BVSR tail. 
             (b) A size 5 ball with 500 g weight and a size 7 ball with 650 g are adopted to generate impact. 
             (c) In each test, a ball is released from the top of a 45°, 1.35 m long ramp to generate identical impact to the tails installed on the quadruped. 
             (d) Velocity responses of the two tails and the quadruped under two impact conditions are plotted. 
        }
        \label{fig:17_TailEffect}
    \end{figure}

    To demonstrate the positive effect of the tail's compliance on maintaining the position and posture stability of a quadruped robot, a rigid prosthetic tail with identical geometric parameters as the BVSR was used as the comparison, as shown in Fig.~\ref{fig:17_TailEffect}(a). The prosthetic tail was fabricated using PLA through FDM printing. The two tails were installed on the quadruped in respective tests and placed at the bottom of the ramp as shown in Fig.~\ref{fig:17_TailEffect}(c). At the beginning of each experiment, a size 5 ball (500 g) or a size 7 ball (650 g), as shown in Fig.~\ref{fig:17_TailEffect}(b), was released from the top of a 1.35 m-long ramp with a 45° incline as shown in Fig.~\ref{fig:17_TailEffect}(c). The ball rolled down and impacted the tail mounted on the quadruped robot. Each test was repeated 3 times. The experimental results are presented in Fig.\ref{fig:17_TailEffect}(d). 

    Under identical impact conditions, the quadruped robot equipped with the BVSR exhibited significantly lower velocity responses than those of the robot with the prosthetic tail. Owing to its compliant structure, the BVSR absorbed and dissipated a portion of the impact energy through deformation, thereby reducing the impact load transmitted to the quadruped torso. Comparisons show that the BVSR tail positively affects the robot's position and posture stability, owing to its compliance. 

\subsubsection{Active Payload}

\begin{figure}[htbp]
        \centering
        \includegraphics[width=0.8\linewidth]{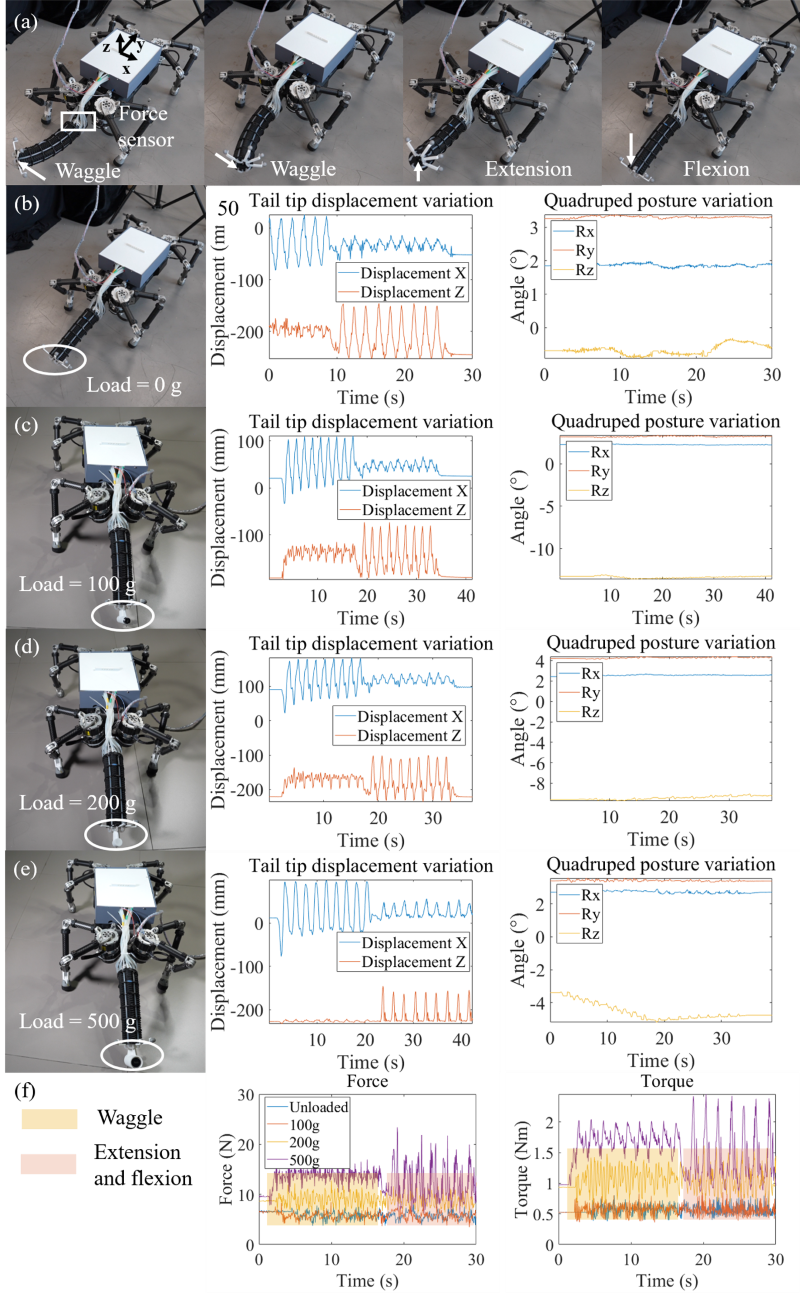}
        \caption{
            \textbf{The effect of the active BVSR motions with payload on the robotic platform.}
            (a) Motion outputs of the BVSR. 
            (b--e) Tail-tip displacements and quadruped posture variations with payloads of 0 g, 100 g, 200 g, and 500 g, respectively.
            }
        \label{fig:18_ActivePayload}
    \end{figure}

    We carried out an active payload experiment to evaluate the influence of BVSR tail motions on the robotic platform as shown in Fig.~\ref{fig:18_ActivePayload}(a). Additional payloads of 0 g, 100 g, 200 g, and 500 g were attached to the tail tip in each test as shown in Figs.~\ref{fig:18_ActivePayload}(b-e), respectively. The tail was then actuated to perform waggle, extension, and flexion motions sequentially, to a tip distance of 100 mm, while the quadruped remained standing still. The tail-tip displacement and the quadruped posture variations were recorded by the motion capture equipment and an IMU on the top of the quadruped torso (JY61, Witmotion), respectively.

    The experimental results show that the quadruped posture remained nearly unchanged under the 0 g, 100 g, and 200 g payload conditions, with minor fluctuations observed in the measured posture angles. When the payload was increased to 500 g, a slight rotation about the Z-axis was observed, while the remaining posture angles stayed essentially constant. Meanwhile, the BVSR successfully executed all prescribed motions under every payload condition. The force and torque generated are recorded by the 6-axis force sensor (KWR46A, KUNWEI) fixed at the root of the tail and connected to the torso, and plotted in Fig.~\ref{fig:18_ActivePayload}(f). With a 500 g payload, the BVSR tail generates a maximum force of 23.6 N and torque of 2.48 N·m at the root. The active payload performance is listed in Table~\ref{tab:4_TailCompare}, showing the BVSR tail with a 500 g tip payload is able to generate torque larger than the state-of-the-art robotic tails, except the much heavier continuum tail (2250 g) \cite{rone2014continuum}.

    These results indicate that the motions of the tail with payloads impose a relatively small influence on the quadruped body. This can be attributed to the substantial mass difference between the tail (283 g) and the quadruped robot (14.5 kg), which limits the tail's inertial influence on the platform. Upon comparison, the tail can generate functionally significant forces to manipulate the posture and trajectory of a light mobile platform (around 700 g), as shown in Fig.~\ref{fig:12_TailCart}.

\subsubsection{Self-contained Actuation}

\begin{figure}[htbp]
        \centering
        \includegraphics[width=1\linewidth]{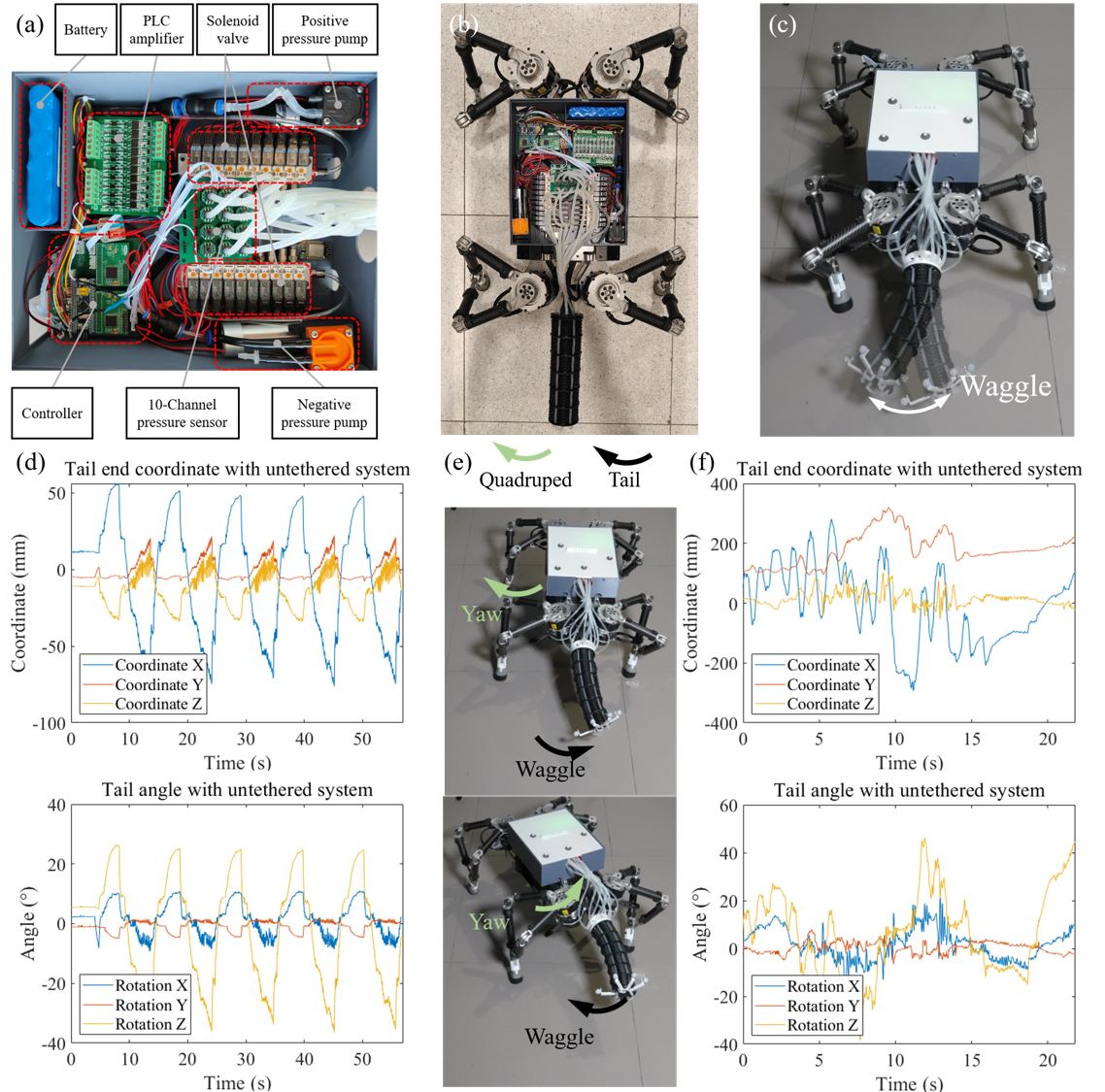}
        \caption{
            \textbf{The untethered self-contained pneumatic actuation system.}
            (a) Prototype demonstration. 
            (b) Quadruped with the self-contained system. 
            (c) Waggle motion with untethered self-contained actuation system, and 
            (d) the corresponding position of the BVSR tail tip was recorded. 
            (e) Coordinated motions of the quadruped robot and the tail with the self-contained system, and 
            (f) the corresponding position of the tail tip were recorded.
          }
        \label{fig:19_UntetheredSyst}
    \end{figure}

   Finally, we developed a self-contained pneumatic actuation system that can be installed on the back of the platform, as illustrated in Fig. ~\ref{fig:19_UntetheredSyst}. The system provides the BVSR tail with pneumatic pressure by integrating a positive-pressure pump (KZP, Kamoer, 24 V, 12 W, 7 L/min, 200 kPa) and a negative-pressure pump (KVP8Plus, Kamoer, 24 V, 9 W, 11 L/min, 0.1kPa). The controller used is STM32F103C8. The control voltage is amplified (Macrowis, NPN) to drive 20-channel solenoid valves (OST Solenoid SY2/2N.C, 24 V, 2.5 W), pressurizing and depressurizing 10 channels. The pressure value of each channel is measured by a pneumatic pressure sensor (XGZP6847A, CFSensor, 0$\sim$200 kPa), collected by a data acquisition module (AD7606, ARMFLY), and sent to the controller via the Serial Peripheral Interface (SPI) protocol. All modules are powered by a battery (24 V, 3,000 mAh) integrated into the system. This pneumatic actuation unit weighs 2.21 kg and provides a maximum input power of 18 W to the BVSR tail. 
   
   With this pneumatic actuation, the BVSR tail was mounted on a robotic platform that formed an untethered robotic system, and output motions as shown in Fig.~\ref{fig:19_UntetheredSyst}(c) (Supplementary Movie S9). Without torso movement, the tail's tip motion was captured by the motion capture equipment and plotted in Fig.~\ref{fig:19_UntetheredSyst}(d), which shows stable coordinates and angular displacements. Coordinated with the torso motions as shown in Fig.~\ref{fig:19_UntetheredSyst}(e) and plotted in Fig.~\ref{fig:19_UntetheredSyst}(f), the tail's tip motion is obviously extended, verifying the tail's potential in constituting a robotic system as a functional limb. This untethered tail-robotic system is a crucial step toward deploying tailed robots for real-world dynamic locomotion tasks.

\section{Discussion}
\label{sec:Discussion}

\subsection{Interpretation of Key Findings}

    Our experiments reveal that the vertebraic design principle directly translates into a $> 200\%$ increase in angular velocity and inertial torque compared to a non-vertebraic soft equivalent. This confirms our primary hypothesis that strategic passive constraint is a viable method for overcoming the intrinsic performance limitations of soft actuators. The high-pressure pulsed actuation strategy, enabled by the vertebra's load-bearing capacity, allows the system to achieve accelerations and generate inertial effects that are functionally significant for dynamic control of the platform of comparable weight, as demonstrated in the mobile cart experiment.

\subsection{The Role of the Vertebra as a Tunable Element}

    The parametric study in Section III-B demonstrates that the vertebra's material properties and geometry are critical tuning parameters, not merely structural reinforcements. This allows the tail's passive stiffness and dynamic response to be customized for specific platforms—a stiffer vertebra (e.g., Nylon, 6 mm) for a heavy quadruped requiring large torques, or a more compliant one (e.g., TPU, 6 mm) for a lighter robot needing a wider range of motion without sacrificing compliance. This tunability is a key advantage of the hybrid vertebraic design, offering application-specific optimization not available in monolithic soft or rigid systems.

\subsection{Comparison to the State of the Art}

    As shown in Table~\ref{tab:4_TailCompare}, the BVSR tail occupies a unique space in the landscape of robotic appendages. Compared to state-of-the-art rigid tails \cite{santiago2016soft, butt2021modeling}, it achieves competitive angular velocities and torques relative to its mass while retaining the inherent safety and compliance of a soft system. Compared to other soft tails, it generates substantially more inertial torque, enabling dynamic regulation of platforms in the 1-5 kg class, a capability previously limited to rigid mechanisms. 
    
    To obtain quantitative proof of the tail's advantage, we calculated the peak power \(P_p \) as the product of the torque and angular velocity \(P_p = \tau \omega\), and obtained the rated power from the driven component of the actuation system; thus, the actuation efficiency is the ratio of peak power over rated power. These new power-related metrics are added to Table~\ref{tab:5_PowerCompare} and compared with existing work that includes such information. Under 6 bar pressure actuation, the tail generates a peak power of approximately 4.5 W, which exceeds that of state-of-the-art robotic tails. When driven by the self-contained actuation system, the peak power is 0.45 W. The efficiency of 1.95\% is greater than the 0.45\% efficiency under laboratory high-pressure actuation and is comparable with the state-of-the-art soft tail. Our work, therefore, successfully bridges the performance-compliance gap, delivering a system that combines the strengths of both paradigms.
    
     \begin{table}[htbp]
        \centering
        \caption{Comparison of Actuation Power and Efficiency}
        \label{tab:5_PowerCompare}
        \resizebox{0.7\columnwidth}{!}{%
            \begin{tabular}{lcccccccc}
                \hline
                \textbf{Robotic Tail} & \textbf{Peak Power (W)} & \textbf{Rated Power (W)} & \textbf{Actuation efficiency (\%)} \\ \hline
                \textbf{BVSR tail} & \textbf{4.5} & \textbf{1000} & \textbf{0.45} \\
                \textbf{BVSR self-contained} & \textbf{ 0.45 } & \textbf{21} & \textbf{1.95} \\
                Lizard tail \cite{changsiu2011a} & - & 3000 & - \\
                Lightweight tail \cite{butt2021modeling} & 0.7 & - & - \\ 
                Prehensile tail \cite{XM2025softtail} & 1.1 & 30 & 3.7 \\ 
                R3RT \cite{saab2019design} & - & 100 & - \\ 
                USRT \cite{rone2018design} & - & 510 & - \\ \hline
            \end{tabular}%
        }
    \end{table}

\subsection{Limitations and Future Directions}

    While our model captures the system's primary dynamics, it does not account for the material's viscoelastic hysteresis, which explains the minor discrepancies between the model and experimental data in Fig.~\ref{fig:9_VerifyKinematics}. According to the active payload experiment in Fig.~\ref{fig:18_ActivePayload} (Supplementary Movie S10), the tail motions and the additional tip payloads impose only a negligible influence on the quadruped body. This is attributed to the substantial mass difference between the tail (283 g) and the quadruped robots (14 kg $\sim$ 14.5 kg), which limits the effects of tail-induced inertial disturbances to the platform. Future work could incorporate more complex constitutive models for higher-fidelity control and will focus on the tail's dynamic effects on the robotic platform to improve coordinated maneuvers. This would be a crucial step toward deploying tailed robots for real-world dynamic locomotion tasks.

\section{Conclusion}
\label{sec:Conclusion}

    In this paper, we introduced a Biomimetic Vertebraic Soft Robotic (BVSR) tail that successfully bridges the gap between the high performance of rigid systems and the inherent safety of soft robotics. Our core contribution is a novel design principle: a compliant body reinforced by a central continuous, constraining vertebra. This bio-inspired architecture enables high-pressure (6 bar) pneumatic actuation, translating directly into superior dynamic capabilities.
    
    We have shown through extensive characterization that this design achieves angular velocities exceeding 670$^{\circ}$/s and generates significant inertial torques (2.48 Nm with 500 g payload), with customization capability in range of motion and inertial performance. These performance metrics are highly competitive with, yet fundamentally safer than, traditional rigid tails. We developed and validated kinematic and dynamic models that predict the system's behavior, providing a foundation for model-based control. Furthermore, a series of functional demonstrations, including repetitive object launching, dynamic assistance, quadruped integration, direct and indirect human interaction, impact resistance, and disturbance absorption, have validated the BVSR tail's practical efficacy and versatility. The developed self-contained pneumatic actuation system also enabled the realization of an untethered tail robot, paving the way for real-world applications.
    
    This work establishes a new, physically grounded paradigm for creating high-performance, compliant robotic appendages. The principles of vertebraic reinforcement and high-pressure pulsed actuation are not limited to tails and could inform the design of next-generation soft manipulators, wearable assistive devices, and dynamic probes for exploring fragile environments. By successfully uniting speed and safety, this research paves the way for robots that can operate dynamically and robustly in the complex, unpredictable human world. Future work will focus on deeper integration with legged platforms, developing closed-loop control algorithms that use the tail for real-time balance recovery and agile maneuvering, fully realizing the potential of this technology.

\section*{Supplementary Materials}
\label{sec:SupMat}

    \begin{itemize}
        \item \textbf{Movie S1}. Motions of the actuator, segment, and BVSR tail.
        \item \textbf{Movie S2}. Dynamic motion repeatability.
        \item \textbf{Movie S3}. Mobile platform assistance.
        \item \textbf{Movie S4}. Integration with a quadruped.
        \item \textbf{Movie S5}. Functional assistance to the quadruped by coordinated motions.
        \item \textbf{Movie S6}. Human-tail direct interaction.
        \item \textbf{Movie S7}. Human-tail indirect interaction.
        \item \textbf{Movie S8}. Impact resistance test of the BVSR tail.
        \item \textbf{Movie S9}. Motion with untethered self-contained actuation system.
        \item \textbf{Movie S10}. Effect of the active BVSR motions on the quadruped.
    \end{itemize}
    
\bibliographystyle{IEEEtran}
\bibliography{References}
\end{document}